\documentclass[10pt,twocolumn,twoside]{IEEEtran}

\usepackage{amsmath,amssymb,amsfonts}
\usepackage{algorithm}
\usepackage{algorithmicx}
\usepackage{algpseudocode}
\usepackage{cite}
\usepackage{caption}
\usepackage{booktabs} 
\usepackage{graphicx}
\usepackage{lipsum} 
\usepackage{tabularx}
\usepackage{textcomp}
\usepackage{bbding}
\usepackage{xcolor}
\usepackage{subfigure}
\usepackage{hyperref}
\DeclareCaptionFormat{myformat}{\fontsize{8}{8}\selectfont#1#2#3}
\def\BibTeX{{\rm B\kern-.05em{\sc i\kern-.025em b}\kern-.08em
    T\kern-.1667em\lower.7ex\hbox{E}\kern-.125emX}}

\usepackage[final]{changes} 
\definechangesauthor[name={Paul}, color=blue]{pz}
\definechangesauthor[name={Gao}, color=red]{gw}

\newcommand{\pza}[1]{\added[id=pz]{#1}}
\newcommand{\pzr}[2]{\replaced[id=pz]{#1}{#2}}

\newcommand{\gwd}[1]{\deleted[id=gw]{#1}}

\usepackage{amsthm}

\newcommand{\setK}{\mathcal{K}}
\newcommand{\setT}{\mathcal{T}}

\newcommand{\bh}{\mathbf{h}}
\newcommand{\bolde}{\mathbf{e}}
\newcommand{\crate}{\hat{r}_{t,k}}

\begin{document}

\title{
Joint Link Adaptation and Device Scheduling Approach for URLLC Industrial IoT Network: A DRL-based Method with Bayesian Optimization
}

\author{Wei Gao, Paul Zheng, Peng Wu, Yulin Hu, and Anke Schmeink
\thanks{Partial material in Section.III, i.e., OLLA assisted execution phase was presented in IEEE Wireless Communications and Networking Conference (WCNC), April 2024, Dubai, United Arab Emirates~\cite{conference}.}
\thanks{The work of P. Zheng and A. Schmeink are supported by BMFTR Germany in the program of “Souverän. Digital. Vernetzt.” Joint Project 6G-ANNA with project identification number 16KISK097.}
\thanks{ W. Gao, P. Wu, and  Y. Hu are with School of Electronic Information, Wuhan University, 430072 Wuhan, China (email: $wei.gao|peng.wu|yulin.hu$@whu.edu.cn). $^*$Y. Hu is the corresponding author.} 
\thanks{P. Zheng, and A. Schmeink is with Chair INDA, RWTH Aachen University, D-52074 Aachen, Germany. (email: $zheng|schmeink$@inda.rwth-aachen.de).}
}

\maketitle

\begin{abstract}
In this article, we consider an industrial internet of things (IIoT) network supporting multi-device dynamic ultra-reliable low-latency communication (URLLC) while the channel state information (CSI) is imperfect. 
A joint link adaptation (LA) and device scheduling (including the order) design is provided, 
aiming at maximizing the total transmission rate under strict block error rate (BLER) constraints.
In particular, a Bayesian optimization (BO) driven Twin Delayed Deep Deterministic Policy Gradient (TD3) method is proposed, which determines the device served order sequence and the corresponding modulation and coding scheme (MCS) adaptively based on the imperfect CSI.
Note that the imperfection of CSI, error sample imbalance in URLLC networks, as well as the parameter sensitivity nature of the TD3 algorithm likely diminish the algorithm's convergence speed and reliability. 
To address such an issue, we proposed a BO based training mechanism for the convergence speed improvement, 
which provides a more reliable learning direction and sample selection method to track the imbalance sample problem.
Via extensive simulations, we show that the proposed algorithm achieves faster convergence and higher sum-rate performance compared to existing solutions.
\end{abstract}


\setlength{\intextsep}{5pt}
\setlength{\textfloatsep}{5pt} 

\section{Introduction}


The Industrial Internet of Things (IIoT) is rapidly advancing toward networks demanding ultra-reliable, low-latency, and massive-device connections. 
Under the fifth-generation (5G) framework, existing Third Generation Partnership Project (3GPP) standards are mainly tailored to periodic, low-frequency transmission characteristics in IIoT networks~\cite{3GPP_R19}, and the link adaptation (LA) and scheduling mechanisms still follow the conventional cellular design paradigm, which is insufficient to satisfy the stringent requirements of IIoT networks in terms of adaptation to highly dynamic environments and deterministic latency guarantees.
Specifically, current LA techniques rely on predefined parameter settings. 
On the one hand, a time-slotted transmission scheme is utilized in \cite{3gpp_AIIOT}, where scheduling cycles and channel hopping patterns work with predefined rules. 
On the other hand, as specified in~\cite{3gpp_fix_CQI_feedback}, the channel quality indicator (CQI) uses periodic feedback intervals, while the blocklength employed for the signal transmission is typically significantly shorter than the CQI feedback period~\cite{CQI_MCS2}. 
This implies that the CSI characterized by CQI, which underpins LA decisions, is not only quantized but also outdated.
Subsequently, the outdated CQI and predefined rules lead to a mismatch between the MCS and the actual channel, resulting in communications unreliability~\cite{high-reliability_in_URLLC}.

For the massive-device connections requirement in IIoT networks, the limited spectrum resources make it difficult to support simultaneous access and guarantee high-reliability, low-latency transmission for all devices~\cite{IIoT,IIoT1}. 
To cope with this limitation, existing studies have proposed several device scheduling methods under constrained spectrum resources by appropriately grouping users or selecting a subset of devices to be served, to improve the transmission rate~\cite{scheduling_wp,Paul_TWC,US_rate_SG,US_rate1_SG}, reduce energy consumption~\cite{scheduling_EE}, and enhance interference suppression~\cite{scheduling_inference}.
However, as revealed in \cite{Performance_Variance}, the random nature of the wireless environment (fading, interference, mobility) results in substantial fluctuations in network performance metrics (spectral efficiency, channel gain, etc.) between consecutive environment features (time slot, location, etc.). 
In addition, the coexistence of heterogeneous services and the dynamic behavior of devices in IIoT networks leads to highly diverse channel-gain variations across devices.
Consequently, the service order among device clusters becomes a critical factor for enabling URLLC transmissions for a large number of devices.
Nevertheless, the aforementioned studies, as well as our previous work~\cite{conference}, have not considered a joint design of LA and device service ordering.

Based on the above analysis, a joint LA and device scheduling to guarantee ultra-reliable and low-latency transmissions for multi-device conditions is urged to be investigated.
This can be easily achieved by using the exhaustive search method~\cite{scheduling_gyf} and the alternating optimization (AO) method~\cite{scheduling_inference2} if the current and future CSI of all users are perfectly known (e.g., under perfect channel estimation and quasi-static channels).
However, the stringent URLLC requirements and the dynamic behavior of devices in the IIoT network make the CSI highly time-varying and difficult to estimate accurately, and the quasi-static CSI assumption no longer holds~\cite{Outdated_CSI_assump}. Thus, the CSI of future time slots is unavailable, making conventional analytically tractable optimization approaches inapplicable.
Recently, deep reinforcement learning (DRL) technique has been widely used in solving complex decision-making problems due to its adaptive and self-regulating capabilities~\cite{RL,DRL_SG,DRL_SG1,DRL_SG2}.
Several DRL-based methods have been proposed for wireless systems design, e.g., deep Q-network (DQN)~\cite{L_DQN,DQN_SG}, deterministic policy gradient (DDPG)~\cite{DDPG_SG}, proximal policy optimization (PPO)~\cite{CMAB,PPO_SG}, and the twin delayed deep deterministic policy gradient (TD3) based MCS selected algorithm~\cite{TD3_LA}, demonstrating the ability of DRL for solving complex problems in dynamic wireless environment.

However, traditional DRL methods are highly sensitive to hyperparameter configuration, and the stringent URLLC requirements further induce a severe imbalance between acknowledgment (ACK) and negative acknowledgment (NACK) feedback samples, which can result in prolonged convergence time, performance fluctuations, or even divergence. The long time required for the learned model to become effective in practice is detrimental to maintaining transmission reliability and improving transmission rates.
To alleviate such limitations, several works combined with convex optimization to improve the convergence speed and reliability of the DRL~\cite{optimization_based_DRL2,convex_with_DRL,DRL_with_LS}.
Specifically, the authors in~\cite{optimization_based_DRL2} introduce the interior-point algorithm and successive convex approximation (SCA) to provide the lower bound for the proposed method to facilitate the DRL model training. 
The authors in~\cite{convex_with_DRL} proposed a block coordinate descent (BCD) based method, where first decompose the original problem into two sub-problems for problem simplicity, and then, introduce the DRL and Lagrangian duality method to solve the sub-problems separately.
Similarly, the authors in~\cite{DRL_with_LS} utilized the least squares method to simplify the original problem, then used the DDPG method to solve the simplified problem.
However, existing convex optimization method driven DRL approaches face two critical challenges. 
First, in the training phase, DRL employs neural networks to construct a black-box mapping between the objective function and the optimization variables, whereas convex optimization-based methods typically require a closed-form expression of the objective function. 
As a result, convex optimization-based methods are ill-suited to characterizing the internal black-box mapping of DRL models and thus cannot be directly exploited to enhance the performance of the DRL training process.
Second, if a convex optimization method is employed in the testing phase, its performance is inherently limited by imperfect CSI feedback, and the high computational complexity of the convex optimization procedure makes it difficult to meet the stringent low-latency requirements of IIoT networks.
To this end, we introduce Bayesian optimization (BO)~\cite{BO}, a representative black-box optimization framework, to capture the internal black-box mapping between the inputs and outputs of the neural networks in DRL.
The BO method can explicitly model predictive uncertainty and, through its acquisition function, adaptively balance exploration and exploitation, achieving high sample efficiency and robustness in noisy and small-sample environments~\cite{BOLA,BO_URLLC,BO_UAV}.
Motivated by these properties, we integrate BO with DRL to accelerate convergence and enhance the stability of the DRL algorithm.

As described previously, in practical dynamic IIoT networks, LA design remains challenging due to outdated and quantized CSI and strict block error rate (BLER) constraints. 
Moreover, substantial channel variations amplify the impact of device scheduling sequences on overall network performance.
To address these challenges, we propose a BO driven DRL based framework aiming to maximize the transmission rate while ensuring adherence to stringent reliability requirements within fixed delay in dynamic multi-device URLLC networks.
Specifically, we adopt the TD3 algorithm, which is widely recognized for its effectiveness in dynamic environments, to determine device scheduling and MCS. 
To alleviate the long convergence time typically observed in DRL training, we further incorporate BO into the training phase aimed at accelerating convergence. Subsequently, the Outer Loop Link Adaptation (OLLA) based execution phase is introduced to enhance the reliability of the learned DRL policy.
The key contributions of this work are summarized as follows:
\begin{itemize}
    \item Considering that the perfect CSI in dynamic IIoT is unavailable, along with the significant variations in wireless channels driven by device mobility and environmental randomness, we proposed a DRL-based method with a novel training phase. This method aims to select the serving device order and the corresponding appropriate MCS with outdated and quantized CSI, to maximize the coding rate while satisfying BLER requirement. 
    \item We propose a BO based training mechanism. Specifically, a BO and modified exponential-weight algorithm for exploration and exploitation~(EXP3) based optimization module is proposed to improve the convergence speed by providing a more stable action selection strategy, allowing the model to learn quickly. 
    Secondly, the sample selection method is provided to tackle the training issue due to the imbalanced samples between ACK and NACK.
    \item Through extensive simulations, we demonstrate that our proposed algorithm outperforms existing methods regarding coding rate performance while meeting the strict expected BLER requirement.
\end{itemize}

The remainder of this paper is organized as follows: 
In Section~II, we present the network model.
Section~III introduces the BO assisted DRL Framework. 
The proposed BO assisted TD3 based link adaptation and device scheduling method are developed in Section~IV.
We provide simulation results in Section~V and conclude the paper in Section~VI.

\section{Related Work}
In this section, we first comprehensively survey related work on LA and device scheduling technologies. 
Subsequently, we delve into an exploration of artificial intelligence (AI) technology based algorithms for dynamic LA and device scheduling, along with our motivations to enhance their efficiency.
A comparative summary is presented in Table~\ref{tab:example}.
\begin{table*}[h]
\centering
\caption{Comparison of proposed work with state of art approaches}
\begin{tabular}{|c|c|c|c|c|c|} 
\hline 
~ & Link adaptation & Device Scheduling & Device Service Order & CSI & URLLC 
\\ \hline 
[56]
& \Checkmark & \XSolidBrush & \XSolidBrush & Perfect& \XSolidBrush 
\\ \hline 
[33]
& \Checkmark & \XSolidBrush & \XSolidBrush & Quantized & \XSolidBrush 
\\ \hline 
[31],[43]
& \Checkmark & \XSolidBrush & \XSolidBrush & Perfect& \Checkmark 
\\ \hline 
[1],[13]
& \Checkmark & \XSolidBrush & \XSolidBrush & Imperfect & \Checkmark
\\ \hline 
[23]
& \XSolidBrush & \Checkmark & \XSolidBrush & Perfect & \XSolidBrush
\\ \hline 
[16],[17]
& \XSolidBrush & \Checkmark &  \Checkmark & Perfect & \XSolidBrush
\\ \hline 
[74]
& \XSolidBrush & \Checkmark & \XSolidBrush & Perfect & \XSolidBrush
\\ \hline 
[26]
& \Checkmark & \Checkmark & \XSolidBrush & Imperfect & \XSolidBrush
\\ \hline 
[71]
& \XSolidBrush & \Checkmark & \XSolidBrush & Imperfect & \Checkmark
\\ \hline 
This work 
& \Checkmark & \Checkmark & \Checkmark & Imperfect & \Checkmark 
\\ \hline 
\end{tabular}
\label{tab:example}
\end{table*}

\subsection{Optimization based link adaptation and device scheduling}
To meet the stringent high-reliability and low-latency communication requirements in IIoT networks, extensive research on LA and device scheduling has produced a series of important results.
Regarding LA, existing studies selected MCS mainly according to the quantized CSI feedback from device terminals~\cite{LA,LA1}.
To enhance link reliability, the authors in~\cite{conservative_LA} propose a conservative LA algorithm that selects the MCS based on worst-case empirical channel prediction.
In~\cite{filtered_LA}, filtered interference information is incorporated into the CQI measurement to improve the accuracy of channel-state matching.
For open-loop LA, the authors in~\cite{OLLA_LA} adjust the transmission strategy based on CQI feedback, but its performance is highly sensitive to errors in the ACK signaling.
For URLLC and enhanced mobile broadband (eMBB) multiplexing networks~\cite{URLLC_eMBB}, a CQI-based LA method is proposed. 
However, the performance of the method degrades significantly when CQI updates are outdated, making it unsuitable for large-scale device network.
To further tackle interference, the interference is treated as the main performance-limiting factor, and its temporal correlation is used to predict the CSI in~\cite{interference_LA1,interference_LA2}. However, such approaches struggle to track the fast channel variations induced by high device mobility.
In dynamic environments, the authors in~\cite{dynamic_LA} jointly account for signal-quality fluctuations due to device mobility, interference-power variations, and the impact of multi-antenna beamforming, and introduce a backoff mechanism based on CSI distribution quantiles to improve system robustness.

For the device scheduling, conventional methods often adopt greedy strategies to maximize the instantaneous transmission rate~\cite{US_rate}, which in general only achieve suboptimal performance.
To fully exploit multi-device scheduling gains, a variety of optimization-based schemes have been proposed.
In~\cite{ordered_scheduling}, an ordered scheduling mechanism for multi-cell coordination is proposed, where the scheduling order is determined to maximize the minimum transmission capacity among users.
For resource limitation scenarios, the authors in \cite{WPT_MEC_US} investigate a wireless power transform network, where random device arrivals and task execution are taken into account in the allocation of scheduling time slots and computing resources for each device. By formulating the optimization problem as a mixed-integer nonlinear program and applying a generalized Benders decomposition method to decompose the original problem into multiple subproblems that are solved iteratively, effectively reduced the overall system energy consumption.
To reduce signaling overhead, the authors in~\cite{Multicell_US} design a distributed multi-device scheduling strategy based on local information to improve the multi-cell sum rate.
In \cite{yuan_US}, massive URLLC device-access networks are considered, which combine broadcast and TDMA access mechanisms, first clustering devices and analyzing the intrinsic relationship between the clustering structure and resource allocation. 
Following a successive convex approximation approach is proposed to jointly optimize the blocklength and transmit power, thereby achieving a substantial reduction in the maximum error probability under high connection density.
To address the stringent delay requirements of URLLC services, the theoretical analysis of the optimal scheduling policy and its performance upper bound for maximizing the number of successfully delivered frames within a given transmission period under hard deadline constraints is provided~\cite{Online_US}. Based on such analysis, a low-complexity online multi-device scheduling algorithm is proposed, which maintains a high on-time delivery probability even in large-scale device networks.
However, all of the above works assume that the instantaneous CSI is perfect and fully available. To cope with imperfect CSI, in \cite{scheduling_inference,scheduling_inference2}, a device scheduling scheme based on statistical CSI is proposed.
The large-scale MIMO downlink networks with channel estimation errors are considered in \cite{scheduling_inference}, which exploits statistical CSI to perform device scheduling and grouping, and subsequently conducts channel estimation for the selected devices to acquire their instantaneous CSI. Jointly accounting for both estimation overhead and errors in the precoder design, enhancing the overall system spectral efficiency.
Furthermore, for complex urban street deployment scenarios, the authors in \cite{scheduling_inference2} account for link blockages and CSI uncertainty caused by device mobility, and introduce reconfigurable intelligent surface (RIS) technology. By leveraging binary-variable relaxation, AO, and SCA to alternately design device scheduling, precoding, and RIS phase shifts, the proposed scheme increases the number of admissible devices while satisfying their individual rate requirements.
In cell-free systems, the semi-definite relaxation and successive convex approximation are utilized in~\cite{cell_free_US} to jointly solve the device grouping and scheduling problem.

\subsection{Machine Learning based link adaptation and device scheduling}
Although optimization-driven scheduling approaches can obtain the optimal solution in principle, however, these methods typically rely on a large number of iterations and convex optimization solvers, which incur a substantial computational delay and make it difficult to satisfy the stringent millisecond-level latency requirements of URLLC services.
In addition, these methods commonly rely on the assumption of instantaneous and perfect CSI, whereas the available CSI is often outdated and imperfect in practical URLLC IIoT networks.
Although some existing works resort to using statistical CSI to design fixed scheduling policies, they still cannot dynamically adapt the scheduling decisions to real-time, imperfect CSI, which substantially limits their applicability in genuinely dynamic environments.
To cope with imperfect CSI, machine learning (ML) has attracted considerable attention in recent years~\cite{AI_WJC,AI_WJC1,MCS_MAB_rate,MCS_MAB_EE,TS_LA,OLLA_MAB,US_DNN}, and has been widely used for LA schemes design.
Within the multi-armed bandit (MAB) framework, the authors in~\cite{MCS_MAB_rate} and~\cite{MCS_MAB_EE} optimize MCS selection to maximize the data rate and energy efficiency.
In~\cite{TS_LA}, a latent Thompson sampling (TS) algorithm is proposed, which enables tuning-free LA.
In~\cite{OLLA_MAB}, the OLLA parameter update process is modeled as an MAB problem, where arm selection is exploited to compensate for MCS bias.
In~\cite{US_DNN}, the authors attempt to formulate device scheduling as a combinatorial optimization problem and employ neural networks to directly capture the map function between the CSI and the scheduled device set.
For URLLC–eMBB multiplexing scenarios, the authors in \cite{BO_URLLC} analytically derive the steady-state distribution of the URLLC queue and formulate the joint utility maximization problem as a stochastic black-box optimization task. A Gaussian process based BO framework is then introduced to optimize the scheduling policy and resource allocation, thereby significantly improving transmission reliability and overall system throughput while satisfying the queueing-delay requirements.
However, such methods face fundamental scalability and practicality challenges. The non-stationary nature of wireless channels often leads to a distribution mismatch between the training and deployment phases, thereby undermining the robustness and generalization capability of the learned models.

Moreover, in recent years, BO has not only been directly applied to solving complex problems in dynamic environments, but has also been widely adopted for the dynamic tuning of hyperparameters in deep learning models.
In \cite{EI}, an uplink open-loop power control (OLPC) design problem is taken into account. Specifically, a Gaussian-process-based BO method is proposed to select the resource management control parameters to be tuned (such as the power-control offset and compensation factor in OLPC), thereby rapidly approaching the optimal parameter configuration with only a limited number of experiments.
Treating different network configurations as distinct tasks, the authors in \cite{BO_OLPC} further exploit shared contextual information across tasks (e.g., link-distance features derived from interference graphs) to enable knowledge transfer, and propose a contextual meta optimization framework that combines meta-learning, BO, and multi-armed bandits. This framework attains near-optimal performance on new configurations with substantially fewer performance evaluations, thereby improving the robustness and generality of the OLPC algorithm.
When the objective function exhibits temporal correlation, the closed-form expression is unavailable, and varies non-stationarily over time, classical Gaussian-process regression becomes inadequate due to both its computational complexity and the stationarity assumption imposed on the kernel.
To overcome this limitation, the authors in \cite{sequ_BO} model the non-stationary objective as an infinite mixture of Gaussian processes, and optimize latent variables such as the locally active GP expert and its hyperparameters at each time instant, enabling online modeling and uncertainty quantification of the time-varying function. Based on such a model, an online Gaussian-process bandit optimization procedure is further developed to improve decision-making performance.
These studies collectively demonstrate the effectiveness of BO for algorithmic hyperparameter selection. Motivated by these advances, in this paper, we integrate BO with DRL to mitigate the hyperparameter sensitivity of DRL algorithms and enhance their reliability.

\subsection{DRL based link adaptation and device scheduling}
To address the fundamental scalability and practicality challenges of ML, the DRL has been progressively employed for LA and scheduling decisions due to its strong capability of adapting to complex environments.
For the LA, the DRL algorithms such as DQN~\cite{DQN_LA,DQN_LA1} and DDPG~\cite{DDPG_LA,DDPG_LA1} are used to determine the MCS and power allocation in multi-device cellular networks.
In~\cite{DRL_US1}, a comprehensive survey of both conventional and RL-based scheduling methods is presented, and summarizes the advantages and performance gains brought by RL.
In \cite{DQN_LA}, for a multi-device cellular network with time-varying wireless channels and a limited number of available resource blocks (RBs), a DQN based joint MCS and RB allocation scheme is proposed under imperfect CSI feedback. Specifically, an action reward branching DQN with multi-dimensional feedback is employed to perform parallel MCS decisions, while an adaptive deferred-acceptance reliability matching algorithm is incorporated to realize low-complexity RB allocation, thereby minimizing the total number of occupied RBs subject to heterogeneous quality of service (QoS) constraints across users.
In \cite{underwater_LA}, for underwater acoustic communication scenarios where accurate instantaneous CSI is unavailable, a double deep Q-network (DDQN) based LA scheme is developed to adaptively select the transmission frequency, power, and rate to maximize the link energy efficiency under different communication ranges and multi-path conditions. In particular, the scheme adopts a dueling echo state network to reduce the complexity of training, and incorporates meta learning to rapidly adjust the network parameters, enabling the policy to approach the optimal configuration after only a small number of interactions in a new environment.
However, most of these schemes are designed to improve eMBB throughput, which falls short of the much more stringent reliability requirements in URLLC networks. Moreover, these methods' reliance on pre-trained models makes it difficult to accommodate the highly dynamic traffic characteristics of URLLC services.
Targeting the stringent reliability requirements of URLLC, a DQN algorithm with a classified experience replay mechanism for periodic CQI feedback is proposed~\cite{DQN_MCS_URLLC}, aiming to maximize the transmission rate while satisfying the BLER constraint. However, the resulting policy does not strictly satisfy the BLER requirement.
To address this issue, the authors in~\cite{L_DQN} further adopt a constrained $\epsilon$-greedy decision strategy and introduce an enhanced deep neural network (DNN) training procedure to refine the policy, thereby meeting the strict BLER requirement.

For the device scheduling design, the authors in~\cite{DRL_US2} model the large-scale MU-MIMO device scheduling process as a Markov decision process (MDP) and solve it via Q-learning. However, the state space dimensionality of Q-learning based method is high, leading to a complex learning procedure and convergence difficulties.
Thus, the DNN is introduced in~\cite{DRL_MU}, which employs the DDPG method to handle large-scale device scheduling. 
In \cite{embb_urllc_US}, a dynamic uplink multi-device MIMO vehicular network is considered. Under simultaneous access of multiple devices and spatial data streams, the enormous scheduling action space and imperfect CSI make conventional static schedulers incapable of maintaining a high system sum rate. To solve such problem, the authors proposed a PPO based device scheduling strategy to jointly design the number of simultaneously served users and their access order, improving the overall throughput under various channel conditions and imperfect CSI.
Furthermore, for heterogeneous service scenarios where eMBB and URLLC traffic coexist, the authors in \cite{server_order_US_drl} take into account the highly heterogeneous delay tolerances and priorities of devices, as well as the random nature of the wireless channels, and propose a distributed DDPG based device scheduling algorithm. By combining Deep Sets with a dueling network architecture, improving the overall energy efficiency.
In \cite{space_ground_US}, a space–ground downlink massive MIMO communication scenario is investigated, where the extremely long propagation distance implies that the ground base station can only acquire outdated CSI, and both the transmit energy budget and the number of available antennas are limited. The device scheduling problem is formulated as an MDP, and a novel DRL algorithm is proposed to learn the optimal device association and antenna selection policy, thereby improving the system sum rate in the presence of imperfect CSI.
The authors in~\cite{DDQN_US} further extend the framework to multi-cell multi-device network and employ a distributed DDQN architecture to jointly design device scheduling and beamforming.

Most of the above studies treat LA and device scheduling as separate problems, without fully accounting for the joint optimization in IIoT networks.
Although DRL based algorithms exhibit considerable potential, they are highly sensitive to hyperparameter settings and, in URLLC networks with imbalanced ACK/NACK sample distributions, are prone to convergence difficulties, which limit their effectiveness for deployment in IIoT networks.



\section{Systems Model}
\begin{figure}[t]
    \centering
    \includegraphics[width=.94\linewidth,trim= 20 20 20 20]{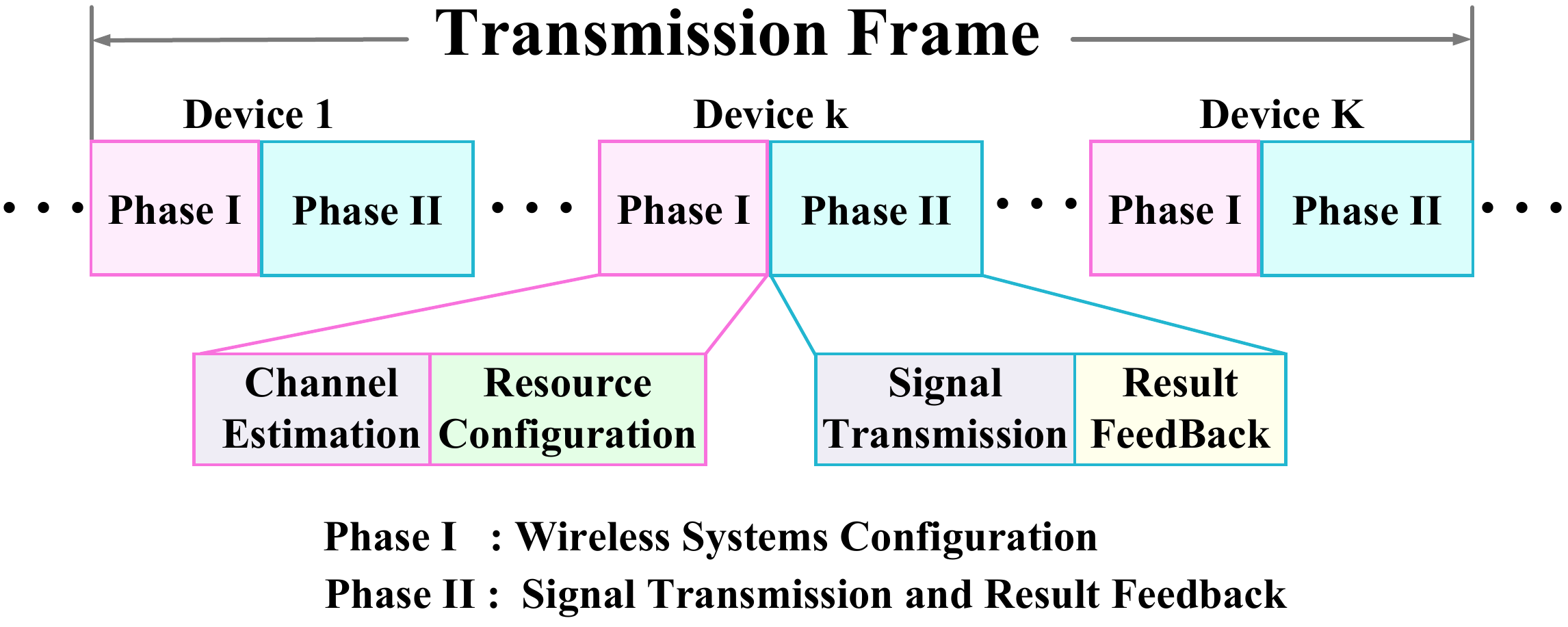}
    \caption{The TDMA based transmission frame structure for the IIoT networks.}
    \label{fig:network_model}
\end{figure}

We consider a downlink multi-device IIoT network, where a central controller with $N$ antennas transmits command signals with short packet size to $K$ single-antenna remote devices (RD). Each RD has a different movement status, including stationary, regular movement with constant speed, irregular movement, etc. 
As shown in Fig.\ref{fig:network_model}, we assume the controller serves the devices using the TDMA scheme, i.e., the transmission frame is divided into $\setT$ time slot and transmits command signals to a single RD for each slot.
Each time slot contains two phases: the wireless networks configuration phase and the signal transmission and result feedback phase. 
In the control parameter phase, the controller transmits the pilot signal for CSI estimation and determines the wireless parameters.
In the transmission and feedback phase, the controller transmitted the command signal with finite blocklength code.
Then, the RD performs channel estimation based on the pilot signal and decoding command signal, feedback on the estimation and reliability requirement index results to the controller. 
Next, we introduce each phase in detail.

\subsection{Control Parameter Phase}
In the control parameter phase at time slot $t$, the controller first selects the RD index $\mathbf{u}_t$ from the set $\{0,1,\dots,K\}$, where each element indicates whether the corresponding device is selected (equal to 1 for selected, $0$ otherwise). 
We assume that within a transmission frame, each RD completing its transmission will not be served again until all other RDs have finished their respective transmissions.
After selecting the RD index, the corresponding MCS value $m$ is determined and used to encode the command signal into a short-sized packet. 
Typically, both device scheduling and MCS selection depend on channel coefficients fed back from RDs. However, due to the small packet size of the command signal, the data transmission duration is often significantly shorter than the time required for accurate channel estimation. Consequently, the channel coefficients frequently become outdated.
For RD $k$ in time slot $t$, we assume controller utilizes the Gauss-Markov model to estimate instantaneous channel coefficients $\bh_{t,k}$ \cite{channel_relation}
\begin{equation}
    \begin{aligned}
       \bh_{t,k} = \rho \hat{\bh}_{t-1,k} + \sqrt{1-\rho^2} \bolde(t),
    \end{aligned}
    \label{eg:channel_relate}
\end{equation}
where $\rho$ is correlation coefficients, and $\mathbf{e}_k$ is a complex Gaussian random variable as $\mathbf{e}(t) \sim \mathcal{C N}\left(0, \mathbf{I}_L\right)$. The zero-forcing~(ZF) beamforming is utilized to eliminate the interference, as $\mathbf{f}_{t,k} = \mathbf{h}^{\mathbf{H}}_{t,k}\left(\mathbf{h}_{t,k} \mathbf{h}^{\mathbf{H}}_{t,k}\right)^{-1}$.

Furthermore, the selected MCS directly impacts both the coding rate and the BLER. According to the discrete mapping defined in 3GPP~\cite[Table 5.1.3.1-3]{MCS}, higher MCS indices correspond to increased coding rates, enabling more information bits per symbol. However, this also leads to elevated BLER, as characterized in~\cite{Polyanskiy_URLLC}. The BLER can be modeled as:
\begin{equation}
    \begin{aligned}
          \epsilon_k \left(\gamma_{t,k}, \crate \right) = \mathcal{Q}\left ( \frac{\text{log}_2\left ( 1+\gamma_{t,k} \right ) - \crate}{\sqrt{W_{t,k}/m}} \right ),
    \end{aligned}
    \label{eg:BLER}
\end{equation}
where $W_{t,k} = 1 - \left ( 1+\gamma_{t,k} \right )^{-2}$ is the channel dispersion, and $\mathcal{Q}(x)=\int_x^{\infty} \frac{1}{\sqrt{2 \pi}} e^{-\frac{t^2}{2}} dt$ is the Gaussian $Q$-function; $ \gamma_{t,k} = \frac{p\left |  \mathbf{h}_{t,k} \right |^{2} }{\sigma^{2}} $ is the signal-to-noise ratio (SNR) and $\left | \cdot\right|$ denotes the Euclidean norm of the vector, $\sigma^{2}$ is the additive white Gaussian noise power, $p$ is the transmit power, $m$ is the blocklength. $\crate$ denotes the coding rate that is related to the MCS value. 
As shown in \eqref{eg:BLER}, the BLER $\epsilon_k$ increase when $\crate$ increases, showing the trade-off relationship between the transmission rate and reliability.
After $(\setT \cdot K)$ time slots, the average total transmission rate of the networks is given as $\hat{r}_\text{sum}=\frac{1}{\setT}\sum_{t\in \setT}\sum_{k\in K} u_{t,k}\crate$, where $a_{t,k}$ is an integer variable, equal to 1 indicates that RD $k$ is selected in time slot $t$, and $0$ otherwise. $D_{k,t}$ is the packet size of a single transmission, defined as $D_{k,t} = m_{k,t} \crate$.

\subsection{Transmission and Feedback Phase}
\begin{figure}[t]
    \centering
    \includegraphics[width=0.95\linewidth,trim= 5 0 0 -10]{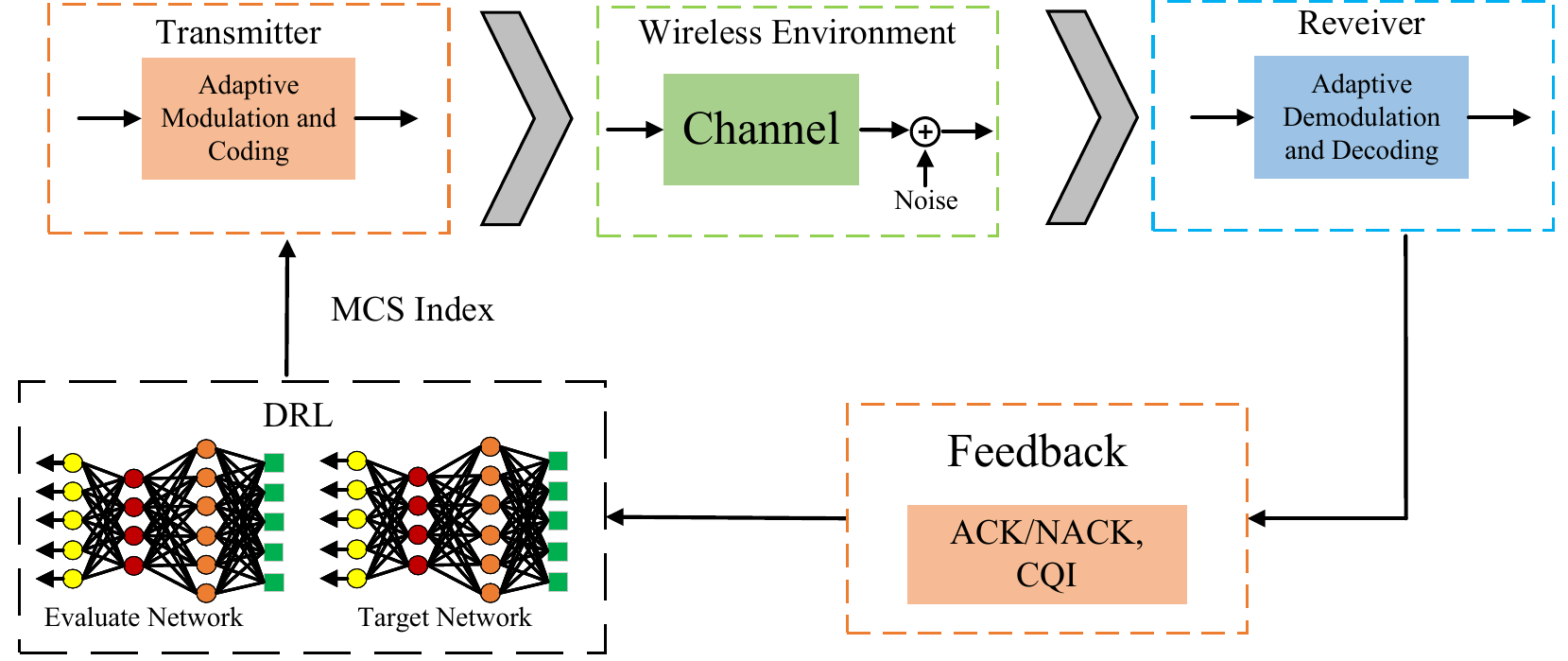}
    \caption{Transmission and Feedback Phase Process Diagram.}
    \label{fig:network_Diagram}
\end{figure}

The Signal transmission and result feedback phase is shown in Fig.\ref{fig:network_Diagram}. 
In such phase, the controller first encodes the command signal into a fixed-size packet based on the selected MCS and transmits it to the RD through a time-varying wireless channel. 
Upon receiving the signal, the RD returns two types of feedback to the BS: the CQI and the reliability requirement index. These feedback components are elaborated in detail below.

Instantaneous CSI characterizes the channel conditions and assists the BS in selecting appropriate MCS values. 
In this paper, we assume perfect CSI acquisition at the receiver RD through existing channel estimation algorithms, as in~\cite{channel_estimation2}. 
However, CSI comprises precise amplitude and phase information across each subcarrier or frequency band, leading to substantial feedback overhead.
Thus, according to the 3GPP standards \cite{3gpp_fix_CQI_feedback,CQI_MCS2}, RD feedbacks CQI for MCS selection.
Specifically, the RD first calculates SNR $\gamma$ based on the estimated perfect CSI and maps SNR to CQI. 
The quantization mapping rule from SNR to CQI is expressed as \cite{CQI}:
\begin{equation}
\text{CQI}= \begin{cases}
    0 & \text {if } \gamma \leq \gamma_{\min}, \\
    \left(N_{CQI}-1\right) & \text {if } \gamma \geq \gamma_{\max },\\
    \left\lfloor\frac{\left(\gamma - \gamma_{\min }\right)\left(N_{CQI}-1\right)}{\gamma_{\max }-\gamma_{\min }}\right\rfloor & \text{otherwise},
\end{cases} 
\label{eg:quantization}
\end{equation}
where $N_{CQI}=2^N$ is the number of CQI index, $N$ is quantization accuracy, $\lfloor\cdot\rfloor$ is the floor function. $\gamma^{max}$ and $\gamma^{min}$ denote the upper bound and lower bound of $\gamma$.
Based on~\eqref{eg:quantization}, the feedback CQI is obtained by quantizing the truly estimated SNR, and thus inevitably introduces quantization errors. 
In general, since the CQI feedback length is predetermined, the mismatch in the reported channel gain caused by this quantization is confined within a bounded region, expressed as 
\begin{equation}
    \mathbf{h}^{true}_{k} = \mathbf{h}^{CQI}_{k} + \Delta \mathbf{h}_{k}, 
    \ \left \| \Delta \mathbf{h}_{k} \right \|_2\leq \varsigma, \ \forall k \in K,
    \label{eg:CSI_uncertainty}
\end{equation}
where $\mathbf{h}^{true}_{k}$ is the channel gain corresponding to the measured SNR, $\mathbf{h}^{CQI}_{k}$ is the channel gain corresponding to the feedback CQI, and $\varsigma$ denotes the error radius.

In URLLC networks, 3GPP standards specify a reliability requirement of $99.9\%$ to support mission-critical industrial automation use cases \cite{reliability_requirement}. 
Typically, receivers acknowledge successful transmissions with ACK and unsuccessful ones with NACK, implying that only one NACK per $1,000$ transmission is acceptable. 
However, directly evaluating reliability through finite transmission counts introduces biases due to arbitrary evaluation boundaries and limitations of the amount of data samples (e.g., initial failures causing non-compliance within $1,000$ transmissions might be compliant over an extended $5,000$ transmission interval) and necessitates extensive computational resources. 
To circumvent these issues, we adopt the BLER as a performance metric. 
Specifically, the RD generates ACK/NACK feedback by comparing BLER computed via \eqref{eg:BLER} against a predefined threshold $\epsilon_{\max}$, i.e., ACK if $\epsilon < \epsilon_{\max}$ and NACK otherwise.

\subsection{Problem Statement}
In IIoT networks, the ultra-high reliability and low latency of single link can be achieved by decreasing the coding rate, resulting in the pronounced loss in spectral efficiency and a drastic increase in the required time-frequency resources.
However, in multi-device networks with limited spectrum resources, the overall service capability of the network is fundamentally constrained by the spectral efficiency.
Specifically, low spectral efficiency implies a smaller packet arrival rate under given bandwidth and latency resources, which in turn leads to longer queueing delays and higher packet loss probabilities, thereby preventing the network from simultaneously guaranteeing high reliability and low latency for a large population of devices.
To address this issue, we aimed to jointly optimize the device serving order and the corresponding MCS level to maximize the network sum rate while maintaining transmission reliability under fixed-blocklength transmission.
Based on the above analysis, the optimization problem can be expressed as: 
\begin{subequations}
	\begin{eqnarray}
         ({\rm OP}):&\max\limits_{\mathbf{u}, \mathbf{\hat{r}}} \, & \frac{1}{\setT}\sum_{t \in \setT}\sum_{k\in \setK} u_{t,k}\crate \\
            &\rm{s.t.:} \!\!\!\!& \epsilon_{t,k} \leq \epsilon_{max} ,\forall t\in \setT, \forall k\in K.\label{con:bler}\\
            &\!\!\!\! & u_{t,k} \in \{0,1\} ,\forall t\in \setT, \forall k\in K, \label{con:device^index}\\
            &\!\!\!\! & \sum_{k\in \setK}u_{t,k} = 1 ,\forall t\in \setT, \label{con:scheduling_device}\\
            &\!\!\!\! & \sum_{t\in \setT}u_{t,k} = 1 ,\forall k\in \setK \label{con:scheduling_time}.
	\end{eqnarray}
 \label{eg:OP}
\end{subequations}
The constraint \eqref{con:bler} denotes the BLER requirement.
The constraints \eqref{con:device^index} denotes that the device selection indicator.
The constraint \eqref{con:scheduling_device} and \eqref{con:scheduling_time} denote that only one RD is selected in each time slot, and no RD is selected more than once.
The problem $(\text{OP})$ is difficult to solve due to the following reasons. 
First, as shown in \eqref{con:device^index}, the RD indicator is an integer variable, which makes problem $(\text{OP})$ a mix\pza{ed}-integer programming problem. 
Second, the perfect CSI is not available due to the CSI uncertainty in~\eqref{eg:CSI_uncertainty}, making the problem non-convex and can not be solved by using convex optimization method.
To solve problem $(\text{OP})$, a novel DRL-based method is introduced in the next section.

\section{BO Assisted Deep Reinforcement Learning Framework}
To address the aforementioned challenges, this section introduces a BO-assisted DRL framework. 
Specifically, an OLLA based execution phase is introduced to enhance transmission reliability, while a BO-assisted training phase accelerates convergence. 
In the following, we first present preliminary conventional DRL methods to highlight their limitations. Subsequently, the OLLA-based execution phase and BO-based training phase are described in detail.

\vspace{-0.4cm}
\subsection{Preliminary of Conventional DRL Method}\label{sec:DRL_pre}

The conventional DRL approach is a classic model-free method that aims to find an optimal mapping function $\pi^*$ between states $\mathbf{s}$ and actions $\mathbf{a}$. 
Specifically, at each time slot $t$, the agent\footnote{For simplicity, in the subsequent text, we will refer to the DRL method as the "agent".} constructs the state $\mathbf{s}_{t}$ and selects an action $\mathbf{a}_{t}$ based on $\pi^*$. 
Upon executing this action, the agent receives an immediate reward $R_t$, reflecting the action's effectiveness. 
The function $\pi^*$ is iteratively updated based on these rewards, aiming to optimize RD selection and MCS assignment. This process continues until convergence to the optimal policy that maximizes cumulative rewards.

To address problem $(\text{OP})$, we adopt the TD3 algorithm, a robust DRL method built upon the actor-critic framework with neural networks. 
TD3 comprises actor and critic networks, each consisting of evaluation and target networks. 
The actor selects deterministic actions based on the current state, and actor's evaluation network parameters $\phi$ are updated using the deterministic policy gradient (DPG) algorithm:
\begin{equation}
    \begin{aligned}
        \nabla_{\phi} J(\phi) = \mathbb{E}_{\mathbf{s}_t \sim \mathcal{D}}\left[\nabla_{\phi}Q_{\theta}(\mathbf{s}_t,\pi_{\phi}(\mathbf{s}_t))\right],
    \end{aligned}
    \label{eg:actor_updated}
\end{equation}
where $Q_{\theta}$ denotes the critic network with parameters $\theta$, and $\mathcal{D}$ is the replay buffer.
The critic network evaluates actions by estimating the Q-value, representing the expected cumulative future reward associated with a given action. 
Critic's evaluation network parameters $\theta$ are updated by minimizing the loss function via gradient descent:
\begin{equation}
    \begin{aligned}
            L(\theta) = \mathbb{E}_{(\mathbf{s}_t,\mathbf{a}_t)\sim \mathcal{D}}\left[\left(Q_{\theta}(\mathbf{s}_t,\mathbf{a}_t)-y_t\right)^2\right],
    \end{aligned}
    \label{eg:critic_loss}
\end{equation}
and the target value $y_t$ is given as
\begin{equation}
    y_t = r_t + \xi \min_{i=1,2} Q_{\theta'^i}\left(\mathbf{s}_{t+1}, \pi_{\phi'}(\mathbf{s}_{t+1}) \right),
    \label{eg:target_value}
\end{equation}
where $R$ is cumulative reward, $\xi$ is discount factor, $Q\left(\cdot\right)$ is the output function of the network, $\theta/ \theta'$ is the evaluate/target network parameter of critic.
To guarantee training stability, the target network parameters are updated in a soft manner, which is updated for every $T_{u}$ time slot as 
\begin{equation}
    \begin{aligned}
        & \theta_1^{\prime} \leftarrow \tau \theta_1+(1-\tau) \theta_1^{\prime}, \\
        & \theta_2^{\prime} \leftarrow \tau \theta_2+(1-\tau) \theta_2^{\prime}, \\
        & \phi^{\prime} \leftarrow \tau \phi+(1-\tau) \phi^{\prime}.
    \end{aligned}
    \label{eg:target_updated}
\end{equation}
The Polyak averaging parameter $\tau$ controls the correlation level between evaluation and target networks, typically set empirically. 
A larger $\tau$ facilitates rapid adaptation to environmental dynamics but risks instability, while a smaller $\tau$ enhances training stability at the cost of slower convergence. Selecting an appropriate value for $\tau$ thus remains challenging.

From the above analysis, conventional DRL methods rely exclusively on reward feedback without explicit mathematical modeling of the environment. Consequently, extensive manual tuning of hyperparameters is required, resulting in unreliable action selection and prolonged training convergence.
\begin{itemize}
\item 
Unreliability primarily arises from uncertainties in state transitions within dynamic environments and the inherent randomness of the $\varepsilon$-greedy strategy. Consequently, identical policies can still yield varying action selections, causing substantial performance fluctuations.

\item The prolonged convergence time primarily results from two factors. 
First, strict reliability requirements in URLLC limit the occurrence of NACK samples, creating an imbalance between ACK and NACK samples. 
This imbalance restricts effective utilization of available data, slowing convergence. 
Second, DRL's inherent sensitivity to hyper-parameters further delays convergence, exacerbated by the absence of standardized selection criteria.
\end{itemize}

\begin{figure*}[t]
    \centering
    \includegraphics[width=0.95\linewidth, trim = 0 170 0 180]{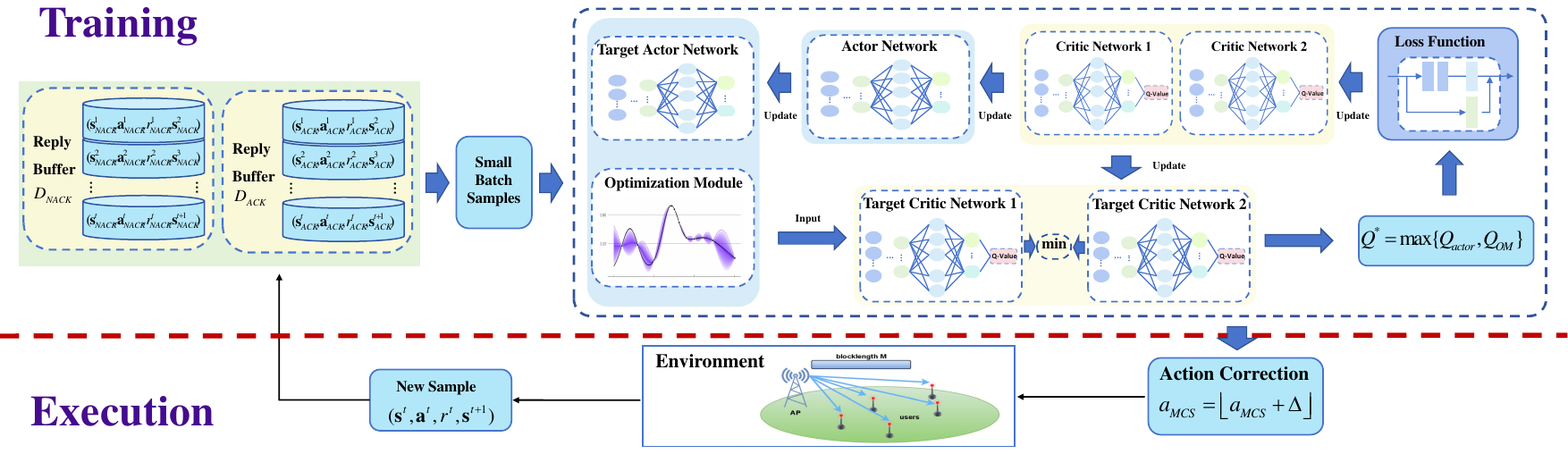}
    \caption{The Flowchart of {BO-TD3} Algorithm.}
    \label{fig:Algorithm}
\end{figure*}

To address the aforementioned problem, we proposed BO assisted TD3 algorithm. The flowchart of the proposed algorithm is shown in Fig.~\ref{fig:Algorithm}. 
The proposed algorithm consists of two phases: execution and training. 
In execution phase, an OLLA-based action correction mechanism is introduced to enhance reliability. 
For training phase, exploiting BO robustness and rapid convergence properties, we propose an GEXP-BO optimization module to expedite convergence. 
This module delivers reliable action selection, thereby guiding the learning process effectively. Additionally, we incorporate a sample selection mechanism to address sample imbalance issues.
Next, we introduce the detail of each phase.

\subsection{OLLA Assisted Execution Phase}
In each time slot, the agent selects the action using the $\varepsilon$-greedy strategy to choose the action, i.e., with a probability of~$\varepsilon$, the model selects an action randomly to prevent the model from getting stuck in a sub-optimal strategy. 
Such randomly selected strategies and the uncertainty of state transitions in dynamic environments can decrease the reliability of networks.
To further improve the reliability of DRL, we utilize the classic OLLA algorithm to modify the selected MCS according to the statistical feedback information from the RD as 
\begin{equation}
    \begin{aligned}
        r_{t,k} = \lfloor r_{t,k}+\Delta_{t,k} \rfloor,
    \end{aligned}
\end{equation}
where $\lfloor \cdot \rfloor$ denotes floor operation and $\Delta_{t,k}$ is the corrective term in time slot $t$ for RD $k$, and updated based on the statistical feedback information as 
\begin{equation}
    \begin{aligned}
        \Delta_{t,k} = \Delta_{t-1,k} + \frac{\pza{\iota}(\epsilon_{\max}-F)}{1-\epsilon_{\max}},
    \end{aligned}
    \label{eg:olla_update}
\end{equation}
where $F=0$ if ACK, and $1$ otherwise. $\iota>0$ is a certain step size. 
The objective of the update in \eqref{eg:olla_update} is as follows: if the BLER exceeds the threshold (i.e., NACK feedback is received), the coding rate is decreased to ensure compliance with the BLER constraint. Otherwise, the coding rate is increased.
It should be emphasized that integrating the OLLA algorithm for action correction is motivated by three primary considerations. 
First, OLLA and DRL share a common objective, where OLLA’s correction terms effectively counter performance degradation due to outdated or quantized CSI, potentially enhancing DRL reliability. 
Second, the operational mechanisms of both algorithms are closely aligned, as each relies on current state information to inform subsequent decisions. Specifically, feedback used to update OLLA correction factors depends directly on previously selected MCS, naturally fitting into the DRL-based Markov Decision Process by incorporating OLLA's corrective terms into the state space.
Finally, OLLA introduces negligible computational overhead since it involves no iterative processes, thus preserving computational efficiency.

\subsection{BO Based Training Phase}\label{sec:training_stage}
In designing the training phase, we focus primarily on addressing slow convergence and sample imbalance issues. To overcome the convergence bottleneck, a BO-assisted optimization module is introduced to guide training more precisely, thereby reducing sensitivity to randomly initialized parameters and hyperparameters and significantly accelerating convergence. 
However, designing this optimization module presents two key challenges. First, direct environment interaction is unavailable during the training phase, precluding real-time evaluation of selected actions. 
Second, obtaining an optimal solution is inherently difficult, as the controller must rely solely on outdated CQI from unserved UEs.

Considering that DRL typically employs a target critic network to evaluate actions, we first adopt the target critic's Q-value output as the evaluation metric. 
However, due to the iterative training and parameter updates in the TD3 algorithm, the resulting Q-value lacks an explicit analytical form, characterizing the network as a black-box optimization problem. 
To solve such difficulty, the BO method is introduced, a well-established and highly efficient approach for black-box optimization problems, leveraging Gaussian process-based surrogate models to efficiently approximate optimal solutions. 

Moreover, since only the selected RD obtains a reward at each time slot, the problem inherently becomes a mixed-integer optimization and multi-armed bandit (MAB) network, which standard BO cannot directly address. 
To overcome this, we introduce an adversarial MAB approach based on the classical EXP3 algorithm. Classical EXP3 cannot be directly applied, as it inherently assumes uniform arm-selection probabilities and a fixed number of arms, conditions unmet in our dynamic network. Thus, we further integrate a gradient bandit mechanism to exploit historical action information, resulting in a novel method termed Gradient EXP3 (GEXP). Consequently, we propose a GEXP-assisted BO algorithm for joint RD and MCS selection, complemented by a tailored sample-selection strategy to effectively mitigate sample imbalance.

\subsubsection{GEXP-assisted BO Based Joint RD and MCS Selection Algorithm}
As discussed previously, since the primary goal during training is selecting actions that maximize Q-values generated by the critic network and assessing actual action performance during the training phase is infeasible, we utilize Q-values from the target critic network as the surrogate objective. The optimization problem is thus formulated as follows:
\vspace{-0.4cm}
\begin{subequations}
	\begin{eqnarray}
         ({\rm P1}):&\max\limits_{\mathbf{a}, \mathbf{r}} \, &
         \frac{1}{\setT}\sum_{t \in \setT}\sum_{t\in \setT} Q\left(a_{k,t}, \crate\right) \\
            &\rm{s.t.:} \!\!\!\!& (\ref{con:bler}),(\ref{con:device^index}), (\ref{con:scheduling_device}), (\ref{con:scheduling_time}).
	\end{eqnarray}
 \label{eg:P_Q_val}
\end{subequations}
where $Q\left(a_{k,t}, \crate\right)$ denotes the Q-value from the target critic network. 
The problem $(\text{P1})$ cannot be solved via classical convex optimization techniques (e.g., SCA or BCD) due to the lack of a closed-form expression for the Q-value. 
Moreover, action selection stability from the target actor network is compromised by DRL parameter sensitivity. 
To overcome these challenges, we propose an GEXP-BO based optimization module, treating each RD as an arm within the GEXP algorithm to accumulate rewards effectively, combined with Gaussian Process (GP) based BO for adaptive and efficient MCS selection.

The core concept behind the GEXP-BO algorithm is the utilization of a surrogate model to approximate the mapping function $\varphi\left (.\right )$ relating the input state $\mathbf{s}$ to its corresponding $Q$-value, denoted as $Q = \varphi(\mathbf{s})$. Using the sample data stored in the agent's replay buffer $\mathcal{D}$, Bayesian updating rules are employed to refine the mean and variance of the surrogate model by computing the posterior distribution $p(\varphi \mid \mathcal{D})$.
Subsequently, based on this posterior distribution, the GEXP algorithm and the BO acquisition function are jointly applied to determine the optimal device index and coding rate.
Specifically, during the $i$-th training iteration, the proposed GEXP-BO optimization module involves the following steps:
\begin{enumerate}
    \item Employ the local dataset $\mathcal{D}$ available at the agent and the BO surrogate model to estimate the posterior probability distribution of the function $p\left(\varphi\left(\mathbf{s}^i\right) \mid \mathcal{D}^i\right)$.\label{MAB_BO_1}
    \item Select the coding rate $r^{i+1}$ utilizing the BO acquisition function, and determine the action $a^{i+1}$ through the GEXP algorithm. Subsequently, the selected rate and MCS values are evaluated using the target critic network to derive the corresponding Q-value.\label{MAB_BO_2}
    \item Update the GEXP weights and enhance the dataset $\mathcal{D}^{i+1}$ accordingly.\label{MAB_BO_3}
\end{enumerate}
Next, we introduce each step in detail.

In Step~\ref{MAB_BO_1}, since the target critic network is incapable of reliably evaluating the effect of actions on future outcomes prior to convergence, $\hat{Q}^{i}$ is introduced as a biased function evaluation, expressed as $\hat{Q}^{i} = \varphi(\mathbf{s}^{i}) + \hat{n}$,
where $\hat{n}$ denotes the bias term. A surrogate model is then employed within the GEXP-BO framework to accurately approximate $\varphi(\cdot)$.

GPs are widely utilized to construct surrogate models in BO, primarily due to their high sample efficiency, non-parametric flexibility, and capability to yield closed-form analytical solutions \cite{GP}. 
Specifically, we consider a zero-mean GP prior for the unknown evaluation function as $\varphi(\mathbf{s})  \sim  \mathcal{GP}(0, \kappa(\mathbf{s}, \mathbf{s}'))$, where $\kappa(\cdot, \cdot)$ denotes the kernel function that quantifies the similarity between input points. 
Given a set of input points $\mathbf{S}^i = [\mathbf{s}^1, \dots, \mathbf{s}^{\mathcal{D}^i}]^\top$, the vector of function evaluations $\varphi^i = [\varphi(\mathbf{s}^1), \dots, \varphi(\mathbf{s}^{\mathcal{D}^i})]^\top$ follows a jointly Gaussian distribution as \begin{equation}
    \begin{aligned}
        p(\varphi_i | \mathbf{S}_i) = \mathcal{N}(\varphi_i; 0, \mathbf{A}_i)
    \end{aligned}
    \label{eg:gp_prior}
\end{equation}
where $\mathbf{A}^i$ is an $\mathcal{D}^i \times \mathcal{D}^i$ covariance matrix, with each element obtained by using the kernel function as $[\mathbf{A}^i]_{\tau, \tau'} = \kappa(\mathbf{s}_\tau, \mathbf{s}_{\tau'})$.

Considering noisy observations, the relationship between the observed values $\mathbf{\hat{Q}}^i = [\hat{Q}_1, \dots, \hat{Q}^i]^\top$ and the true function evaluations $\varphi^i$ can be expressed through a conditional Gaussian likelihood as 
\begin{equation}
    \begin{aligned}
        p(\mathbf{\hat{Q}}_i| \varphi_i, \mathbf{S}_i) = \mathcal{N}(\mathbf{\hat{Q}}_i; \varphi_i, \sigma_o^2 \mathbf{I}_i),
    \end{aligned}
    \label{eg:gp_conditional}
\end{equation}
where $\sigma_o^2$ denotes the observation noise variance, and $\mathbf{I}^i$ represents the identity matrix of size $\mathcal{D}^i$. 
Applying Bayes' rule and combining the prior distributions in \eqref{eg:gp_prior} and likelihood distributions in \eqref{eg:gp_conditional},, the posterior distribution of $\varphi^i$, conditioned on the dataset $\mathcal{D}^i$ is expressed as 
\begin{equation}
    \begin{aligned}
        p(\varphi(\mathbf{s}) | \mathcal{D}) = \mathcal{N}(\varphi(\mathbf{s}); \mu_i(\mathbf{s}), \sigma_i^2(\mathbf{s})).
    \end{aligned}
    \label{eg:gp_posterior}
\end{equation}
The $\mu^i(\mathbf{s})$ and $\sigma^i(\mathbf{s})^2$ denote the mean and variance of the posterior distribution, respectively, given as:
\begin{equation}
    \begin{aligned}
        \mu^i(\mathbf{s}) = \mathbf{A}^i(\mathbf{s})^\top (\mathbf{A}^i + \sigma_o^2 \mathbf{I}^i)^{-1} \mathbf{Q}^i,
    \end{aligned}
    \label{eg:gp_posterior_mean}
\end{equation}
\begin{equation}
    \begin{aligned}
        \sigma^i(\mathbf{s})^2 = \kappa(\mathbf{s}, \mathbf{s}) - \mathbf{A}^i(\mathbf{s})^\top (\mathbf{A}^i + \sigma_o^2 \mathbf{I}^i)^{-1} \mathbf{a}^i(\mathbf{s}),
    \end{aligned}
        \label{eg:gp_posterior_variance}
\end{equation}
where $\mathbf{a}^i(\mathbf{x}) = [\kappa(\mathbf{x}_1, \mathbf{x}), \dots, \kappa(\mathbf{x}^i, \mathbf{x})]^\top$ represents the covariance between the new input state $\mathbf{x}$ and the observation value. 
The posterior distribution of $\varphi(\mathbf{s})$ will be used for the MCS determination in next step. 

In step~\ref{MAB_BO_2}, the core of the GEXP algorithm lies in its exponential weighting mechanism. To this end, we introduce the expert weight vector $\mathbf{w}^{i}$ and preference weight vector $\mathbf{d}^{i}$ to determine RD selection probabilities.
Given the time-varying number of available arms, the selection probability for an unserved RD $k$ is defined as $\hat{p}^i_k = (1 - \zeta) \frac{w_k^{i}}{\sum_{j=1}^K w_j^{i}} + \frac{\zeta}{K+1}$, where $\zeta \in (0, 1]$ is a tunable parameter controlling the exploration-exploitation trade-off. For RDs already served in the current communication round, $\hat{p}^i_k$ is set to zero to ensure the normalization constraint $\sum_{k \in K} \hat{p}^i_k = 1$.
To further incorporate historical knowledge and reward feedback, the preference weight for RD $k$ is defined as $d_k^i = \hat{d}_k^i \cdot v_k^i$, where $v_k^i$ denotes the output of the actor network based on historical environmental inputs, and $\hat{d}_k^i$ is updated based on reward signals from RD selection (as described in Step~\ref{MAB_BO_3}). Combining $\mathbf{d}^i$ and $\hat{\mathbf{p}}^i$, the final RD selection probability is computed via preference-assisted softmax:
\begin{equation}
    p_k^i = \frac{\exp(d_k^i) \cdot \hat{p}_k^i}{\sum_{j=1}^K \exp(d_j^i) \cdot \hat{p}_j^i}.
    \label{eg:RD_pro_vec}
\end{equation}
With the selected RD index $a^{i+1}$, the corresponding MCS level $r^{i+1}$ is determined by maximizing the Expected Improvement (EI) acquisition function~\cite{EI}:
\begin{equation}
    \begin{aligned}
r^{i+1} & =\underset{0< r \leq r_{max}}{\arg \max }  \operatorname{EI}\left(r \mid \mathcal{D}^i, \mathbf{s}^{i+1}\right) \\
& =\underset{0< r \leq r_{max}}{\arg \max }
\left [ \left(\mu^i(r, \mathbf{s}^{i+1})-Q^*\right) \Phi(Z)+\sigma^i(r, \mathbf{s}^{i+1}) \phi(Z) \right]
    \end{aligned}
    \label{eg:EI}
\end{equation}
where $Q^* = \max(\hat{Q}^1, \dots, \hat{Q}^i)$, and $\mu^i$ and $\sigma^i$ denote the posterior mean and variance, respectively. And $Z = \frac{\mu^i(r, \mathbf{s}^{i+1}) - Q^*}{\sigma^i(r, \mathbf{s}^{i+1})}$, with $\Phi(Z)$ and $\phi(Z)$ representing the CDF and PDF of the standard normal distribution. With closed-form expressions of $\mu^i$ and $\sigma^i$ from \eqref{eg:gp_posterior_mean} and \eqref{eg:gp_posterior_variance}, problem \eqref{eg:EI} can be efficiently solved via gradient ascent.
Finally, the selected action $\{a^{i+1}, r^{i+1}\}$ is fed into two target critic networks to evaluate the Q-values, denoted as $Q^{1,i}(a^{i+1}, r^{i+1})$ and $Q^{2,i}(a^{i+1}, r^{i+1})$. To mitigate overestimation, the minimum of the two is used as the final reward for updating the weight parameters, $Q^{i+1} = \min \left\{ Q^{1,i}, Q^{2,i} \right\}$.

In step~\ref{MAB_BO_3}, the GEXP algorithm updates the weight vector based on the observed reward $Q_{i+1}$, thereby progressively increasing the selection probability of actions yielding higher returns. Specifically, GEXP employs importance sampling to construct an unbiased reward estimate as $\hat{\varphi}_{i+1}^m(k) = \frac{Q_{i+1}}{p_k^i \beta}, \quad \forall k \in \{1, \ldots, K\}$, where \( \beta \) is the implicit exploration parameter used to prevent exponential growth in weights and suppress extreme selection probabilities during updates. Accordingly, the weight vector \( \mathbf{w}^i \) is updated as:
\begin{equation}
    w_k^{i+1} = w_k^i \cdot \exp\left( \frac{\gamma \hat{\varphi}_{i+1}^m(k)}{K+1} \right),
    \label{eg:weight_updated}
\end{equation}
where $\gamma$ is the learning rate. Next, we update the preference vector $\hat{\mathbf{d}}$ using a gradient ascent-based rule:
\begin{equation}
\begin{aligned}
    \hat{d}_k^{i+1} &= \hat{d}_k^i + \alpha (Q^{i+1} - \bar{Q}) (1 - p_k^i), && \text{if RD $k$ is selected}, \\
    \hat{d}_k^{i+1} &= \hat{d}_k^i - \alpha (Q^{i+1} - \bar{Q}) p_k^i, && \text{otherwise},
\end{aligned}
\label{eg:preference_updated}
\end{equation}
where $\alpha$ is the step-size parameter and $\bar{Q}$ denotes the running average of historical Q-values.
Subsequently, a new training sample is generated based on the selected action $\{r_{i+1}, a_{i+1}\}$ and the observed reward \( Q_{i+1} \), denoted as \( (\mathbf{X}_{i+1}, Q_{i+1}) \), where \( \mathbf{X}_{i+1} = \{r_{i+1}, \mathbf{s}_{i+1}\} \). The dataset is then updated as $\mathcal{D}_{i+1} = \mathcal{D}^i \cup \left\{ (\mathbf{X}_{i+1}, Q_{i+1}) \right\}$.
In summary, the proposed GEXP-BO joint optimization framework for RD index and MCS selection is detailed in Algorithm~\ref{alg:GEXP-BO}.

\subsubsection{Sample Selection Method}
In URLLC networks, the imbalance between NACK and ACK samples can degrade model performance due to the rarity of negative events. 
To address this, NACK samples are stored in a dedicated replay buffer $\mathcal{D}_{\text{NACK}}$ and periodically sampled for every $T_n$ time slot. 
Considering the temporal dynamics of wireless channels, the correlation among samples diminishes over time. 
Thus, for replay buffer $\mathcal{D}_{\text{ACK}}$, half mini-batches sample ($\frac{\mathcal{D}_B}{2}$) are selected randomly from the replay buffer, and the remaining half is constructed by previous $\frac{\mathcal{D}_B}{2}$ time slot.
When using replay buffer $\mathcal{D}_{\text{NACK}}$, regarding their scarcity and weak temporal correlation, all mini-batch samples ($\mathcal{D}_B$) are selected from $\mathcal{D}_{\text{NACK}}$ randomly.


\section{BO-Assisted TD3 Based Link Adaptation And Device Scheduling Method}
This section provides the overall framework of the proposed BO-assisted TD3-based LA and device scheduling method. 
Specifically, we first introduce the action space, state space, and reward definition in detail. 
Then summarize the overall process of the proposed algorithm for jointly deciding the RD index and corresponding MCS, following the computational complexity discussion. 
 \begin{algorithm}
    \caption{{GEXP-BO} based Algorithm to solve \eqref{eg:P_Q_val}.}
    \begin{algorithmic}[1]
          \State Initialize parameters $\mathcal{D}_0$,  $w_0^k=1, \forall k \in \{1,\cdots,K\}$;
          \For{for time slots $i=0,1,2,...,T$}
             \State Calculate the mean and variance of the posterior distribution according to \eqref{eg:gp_posterior_mean} and \eqref{eg:gp_posterior_variance} with dataset $\mathcal{D}^i$;

             \State Obtain the probability vector $\mathbf{p}^i$, according to \eqref{eg:RD_pro_vec};

             \State Selected the RD index $a^i$ according to $\mathbf{p}^i$;

             \State Determined the MCS value $\crate$ by solving \eqref{eg:EI};

             \State Input $a^i$ and $r^i$ into the two target critic networks to obtain the Q-value, as $Q^i = \min \{Q_{1,i}, Q_{2,i} \}$.

             \State Update the dataset as~$\mathcal{D}_{i+1}=\mathcal{D}^i \cup \left\{\left(\mathbf{X}_{t+1}, Q_{i+1}\right)\right\}$ and weight according to \eqref{eg:weight_updated} and \eqref{eg:preference_updated};
          \EndFor
 \end{algorithmic}
 \label{alg:GEXP-BO}
\end{algorithm}

\subsection{Action-State-Reward Definition}
In this part, we provide the detailed action space, state space, and reward definition.

\subsubsection{Action Space}\label{sec:action_space}
In this work, the action space $\mathcal{A}$ comprises the indices of all unserved RDs and their corresponding candidate MCS values. 
Considering the neural network's fixed output dimensionality, a mask vector $\mathbf{m}t = {m{1,t}, \cdots, m_{K,t}}$ is introduced, where each element represents an RD's availability ($m_{k,t}=1$ if unserved; otherwise, $m_{k,t}=0$). 
Furthermore, due to the temporal correlation of the channel and dependency of optimal MCS selection on instantaneous CSI, the optimal MCS exhibits temporal consistency across successive time slots. 
Exploiting this feature, the network outputs an incremental MCS adjustment $\Delta_{t,k}$, designed to alleviate performance instability caused by outdated CQI. 
Hence, the MCS for RD $k$ at time slot $t$ is determined by $r_{t,k} = r_{t-1,k} + \Delta_{t,k}$. 
Consequently, the action space at time slot $t$ is defined as $\mathcal{A}_t = \left \{ m_{1,t}\mathrm{RD}_{1,t}, m_{2,t}\mathrm{RD}_{2,t},\dots, m_{K,t}\mathrm{RD}_{K,t}, r_{t,k}\right \}$, where $\mathrm{RD}_{k,t}$ denotes the index of the $k$-th RD.

\subsubsection{State Space}
State space $\mathcal{S}$ is constructed by the following two components:
\begin{itemize}
\item Currently available and historical CQI: 
    To mitigate discrepancies between feedback CQI and actual CSI due to delays and quantization errors, we incorporate the current CQI $\mathrm{CQI}t$ into the state space. Exploiting the temporal correlation inherent in wireless channels, we also include CQI values from the preceding $\tau$ time slots, represented as ${\mathrm{CQI}_{t-\tau,1:K}, \dots, \mathrm{CQI}_{t-1,1:K}}$, to enhance the model's ability to capture and adapt to evolving environmental conditions.
\item Feedback and selected action in the last time slot: 
    Incorporating only CQI, even with historical data, into the state space is insufficient for constructing a Markov Decision Process (MDP) in TD3, as CQIs are intrinsic to the wireless environment and independent of the agent's actions. 
    To address this limitation, we augment the state space by including the previous action $a_{t-1}$, receiver feedback $\left(\text{ACK/NACK}\right)_{t-1}$, and the corrective term $\Delta_{t-1}$, which facilitates the formulation of a valid MDP for DRL applications.
\end{itemize}
In summary, the state space of a single RD is expressed as $\mathcal{s}_k=\{\mathrm{CQI}_{k, t-\tau:t},  \left(\text{ACK/NACK}\right)_{t-1}, \mathbf{a}_{t-1}, \mathbf{r}_{t-1}, \Delta_{t-1}\}$.
Therefore, in each time slot, the state space includes the aforementioned information for all RDs. Based on the mask vector, the state space can be represented as $\mathcal{S} = \{m_{1}\mathcal{s}_1, m_{2}\mathcal{s}_2, \cdots, m_{K}\mathcal{s}_K\}$. 


\subsubsection{Reward Function}
The objective of the DRL framework is to maximize the coding rate while satisfying stringent BLER constraints. Thus, we define the reward function for RD $k$ as: 
\begin{equation} 
    r_k = \begin{cases} 
        \beta+ \crate & \text {if success}, \\ 
        \varpi \crate & \text {if fail}, 
    \end{cases} 
\end{equation}
where $\varpi$ represents the proportion of ACK feedback from RD $k$. The introduction of $\varpi$ incentivizes higher MCS selections following successful transmissions, as $\varpi \rightarrow 1$ when transmissions consistently succeed. This design prevents reward collapse upon transmission failures, facilitating continued exploration and exploitation of higher MCS values. Additionally, a reward threshold parameter $\beta>0$ ensures that the reward from a successful transmission (even at lower MCS) always exceeds that from any failed transmissions, i.e., $r^{\text{max}}{\text{fail}}<r^{\text{min}}{\text{success}}$.

\subsection{Overall Algorithm And Complexity Analysis}
In summary, the proposed BO-based TD3 framework is presented in Algorithm~\ref{alg:BO-TD3}.
\gwd{Compared to the traditional TD3 algorithm, the proposed method introduces an additional action correction step during the execution phase and an additional optimization module during the training phase.}
Specifically, the algorithm starts from the execution phase. The agent first selects the RD index and the corresponding MCS, following the MCS correction (Step $3-4$), as $MCS_{(t,k)} = MCS_{(t-1,k)} + \Delta_{t,k}$. 
Next, the agent interacts with the environment to obtain the reward and the next state to construct training samples (Steps $5-8$).
In the training phase, the samples are selected from the replay buffer, following the target actor and Algorithm \ref{alg:GEXP-BO} is used to obtain $a_{\text{ac}}$ and $a_{\text{BO}}$, respectively. 
Then,  $a_{\text{ac}}$ and $a_{\text{BO}}$ are fed into two target critic networks to obtain $Q_{\text{ac1}}$, $Q_{\text{ac2}}$, $Q_{\text{BO}1}$ and $Q_{\text{BO2}}$.
The optimal Q-value $Q^*$ is selected according to $Q^* = \max \{ Q^*_{\text{ac}}, Q^*_{\text{BO}} \}$, where $Q^*_{\text{ac}} = \min\{ Q_{\text{ac1}}, Q_{\text{ac2}} \}$ and $Q^*_{\text{BO}} = \min\{ Q_{\text{BO1}}, Q_{\text{BO2}} \}$ (Steps $9-10$). 
Based on \eqref{eg:actor_updated} and \eqref{eg:critic_loss}, the DPG and SGD algorithms are used to update the parameters of the actor and critic model, respectively (Steps $11$). 
Finally, the target network parameters are updated according to \eqref{eg:target_updated} for every $T_u$ training iteration (Steps $12$). 
\begin{algorithm}
    \caption{{BO-TD3} Algorithm}
    \begin{algorithmic}[1]
          \State Initialize parameters $\theta$,  $\theta'$, $\phi$, $\phi'$, $\Delta$;
          \For{for time slots $t=0,1,2,...,T$}
             \State Selecting the RD index and MCS deviation $\Delta_{t,k}$ based on device's feedback;

             \State Obtain the MCS value by combining the OLLA corrective term;

             \State Transmitting signal by using $MCS_{(t)}$ to obtain the feedback and reward;

             \State Update the OLLA parameter $\Delta_{t,k}$;

             \State Construct the next time-step state $\mathbf{s}_{(t+1)}$;

             \State Store sample data in replay buffer based on feedback;

             \State Sample mini-batch of size $D_{B}$ transitions from the replay buffer;

             \State Utilize the actor network and Algorithm~\ref{alg:GEXP-BO} to obtain the Q-value according to~$Q^* = \max \{ Q^*_{\text{ac}}, Q^*_{\text{BO}} \}$.
            
             \State Compute the loss function in \eqref{eg:critic_loss} and minimize the loss by employing DPG and SGD;
             \State Every $T_u$ time slots update $\omega$ in \eqref{eg:target_updated}.
          \EndFor
 \end{algorithmic}
 \label{alg:BO-TD3}
\end{algorithm}
        
In the final part of this section, we analyze the proposed algorithm's computational complexity. 
The computational complexity of the training phase primarily arises from the Algorithm~\ref{alg:BO-TD3} (BO basd optimization module) and the TD3 algorithm.
The main computational complexity of Algorithm~\ref{alg:BO-TD3} is original from the means and variance calculations in \eqref{eg:gp_posterior_mean} and \eqref{eg:gp_posterior_variance}, leading to a complexity of $\mathcal{O}\left( \left | \mathcal{D}^i \right|^3 \right)$, where $\left | \mathcal{D}^i \right|$ denote the samples number in $i-th$ iterations. The value of $\left | \mathcal{D}^i \right|$ is typically small, often on the order of a few hundred or thousand.
The TD3 algorithm employs one actor framework and two critic frameworks, each comprising primary and corresponding target networks implemented as fully connected neural networks. 
Specifically, the actor network contains an input layer with $K(\tau + 4)$ neurons, an output layer with $K + 1$ neurons, and $L^{a}$ hidden layers with $Z^{a}_l$ neurons in the $l$-th layer. Each critic network consists of an input layer with $K(\tau + 4) + 2$ neurons, an output layer with a single neuron, and $L^{c}$ hidden layers, each containing $Z^{c}_l$ neurons. The target networks share identical architectures with their respective primary actor and critic networks.

During the model training phase, the parameters of all actor and critic networks are updated based on the adaptive moment estimation (Adam) algorithm. Therefore, the total computational complexity of a single parameter update process in Algorithm~\ref{alg:BO-TD3} is given as
\begin{equation}
\begin{aligned}
   \mathcal{O} & \left( \left|\mathcal{D}_i\right|^3 + 2 [ K(\tau+4)Z^{a}_{1} 
   + (K+1)Z^{a}_{L^{a}}  + \sum^{L^{a}_{l+1}}_{l=1} 
   Z^{a}_{l} Z^{a}_{l+1} ] \right.\\ 
   & \left.
   + 4  [(K(\tau+4)+2)Z^{a}_{1} + Z^{c}_{L^{c}} 
   + \sum^{L^{c}_{l+1}}_{l=1} 
   Z^{c}_{l} Z^{c}_{l+1} ] \right),
\end{aligned}
\end{equation}
In the test stage, only the actor network is utilized for action selection, and thus, the computational complexity of the Algorithm~\ref{alg:BO-TD3} is given as $\mathcal{O}\big( K(\tau+4)Z^{a}_1 + (K+1)Z^{a}_{L^{a}} + \sum_{l=1}^{L^{a}_{l+1}} Z^{a}_l Z^{a}_{l+1} \big)$.

\section{Numerical Results}
The simulation setups are described as follows. 
The devices are randomly distributed within a circular area with a radius of \( R \) and move in uniform circular motion with a radius of \( r \) and a speed of \( v \). The value of \( R \) ranges from 8 to 13 meters, \( r \) ranges from 1.5 to 5 meters, and \( v \) ranges from 1.5\,\text{m/s} \, \text{to} \, 2.5\,\text{m/s}, with a 0.1\,s pause after completing one full revolution.
We assume the channel coefficient $\mathbf{h}_k$ between controller and RD $k$ follow Rician fading as in \cite{rician_channel}, where
\begin{equation}
        \mathbf{h}_{k} =  \sqrt{\frac{ \kappa  L(d)}{K+1}} \mathbf{h}^{\mathrm{LoS}} + \sqrt{\frac{L(d)}{K+1}} \mathbf{h}^{\mathrm{NLoS}},
\end{equation}
where $L(d)\!=\!\rho_0\left(\frac{d}{d_0}\right)^{\!\!-\!\alpha}\!$ represents pathloss,
$d$ is the Euclidean distance between the BS and the RD,
$\rho_0$ represents the path loss at the reference distance,
$\alpha$ is the path loss coefficient,
$ \kappa $ is the Rician factor.
$\mathbf{h}^{\mathrm{LoS}}=\left[1,  e^{j\pi \sin (a)}, \ldots, e^{j(M-1) \pi \sin (a)}\right]^T$ represents the line-of-sight (LoS) component with $\sin (a)=\frac{y_{B S}-y_{u}}{\sqrt{\left(x_{B S}-x_{u}\right)^2+\left(y_{B S}-y_{u}\right)^2}}$, $a$ is the angular parameter, and $\mathbf{h}^{\mathrm{NLoS}}$ represents the non-line-of-sight (NLoS)\pza{ component}. 
the NLoS component in $\mathbf{h}_k$ follows CDL-C channel model, which is present in the 3GPP communication protocol \cite{CDL}.

During the TD3 model training process, the exploration rate decreases gradually with the number of time slots. We set ReLU as the activation function and the optimizer as Adam. For GEXP-BO algorithm, the kernel is set as the  Matérn function \cite{gp_kernel}.
TD3 model utilizes ten hidden layers with $600$ neurons, respectively. 
Through numerous simulations, we set $\iota=0.09$, $\tau = 12$,  and $\beta=4$ to achieve better networks performance. The rest of the wireless network and TD3 model parameters are summarized in Table \ref{tab:parameter}.
In addition to our proposed algorithm, the following schemes
are performed for performance comparisons: 
\begin{enumerate}

\item \emph{Ideal}: The controller adopts an exhaustive search algorithm \cite{scheduling_gyf} with perfect CSI to jointly optimize RD serving order and MCS, providing the performance upper bound.

\item \emph{L-DQN}: The layered DQN proposed in \cite{L_DQN} is employed for joint optimization of RD serving order and MCS selection.

\item \emph{BO-CMAB}: This scheme integrates the UCB-based MAB algorithm \cite{CMAB} for RD selection and the BO method \cite{BOLA} for corresponding MCS determination, jointly solving problem $(\text{OP})$.

\item \emph{OLLA-CMAB}: A conventional OLLA algorithm \cite{OLLA} with step size $\iota=0.01$ and convergence threshold $\epsilon_{max}=10^{-3}$ is employed for MCS selection, while the RD index is selected using the MAB algorithm \cite{CMAB}.
\end{enumerate}

\begin{table}[t]
  \centering
  \caption{networks parameters}
  \begin{tabular}{ccc}
    \toprule
    \textbf{Parameter} & \textbf{Description} & \textbf{Value} \\
    \midrule
    $M$ & Antenna number in controller  & 4 \\
    $K$ & RD number  & 4 \\
    $N$ & CQI quantization precision & 4 bit \\
    $\epsilon_{\max}$ &  BLER threshold & $10^{-3}$ \\
    $d_0$ & reference distant & 1\,m \\
    $ \kappa $& Rician factors  & 3\,dB \\
    $B$ & Blocklength & 192 \\
    $\rho_0$ & path loss at the reference distance & -65 \\
    $p$ & transmit power & 35\,dBm \\
    $\sigma$ & noise power density& -105\,dBm \\
    $\beta$ & learning rate & $10^{-3}$ \\
    $\gamma$ & discount factor &  0.99\\
    $T_{u}$ & target update time slots & 400 time slots  \\
    $T_{n}$ & replay buffer change time slots & 5 time slots \\
    $D_{B}$ & batch size & 64 \\
    \bottomrule
  \end{tabular}
  \label{tab:parameter}
\end{table}

\subsection{Convergence Performance}\label{Sec:sim_convgerence}
\begin{figure}[t!]
    \centering
    \includegraphics[width=.9\linewidth, trim = 0 0 0 10]{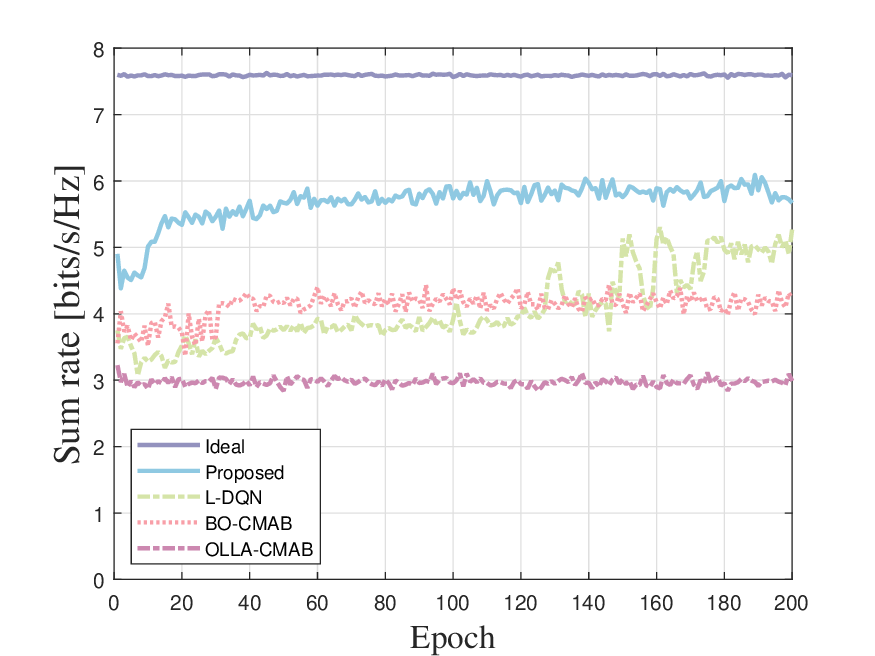}
    \caption{Sum rate performance of different schemes over training epoch.}
    \label{Fig:Training_SP}
\end{figure}
In Fig.\ref{Fig:Training_SP}, we evaluate the sum rate performance versus the training epoch of each scheme, where each epoch consists of $400$ time slots. 
As shown in Fig.\ref{Fig:Training_SP}, the BO-CMAB based scheme achieves faster convergence and exhibits smaller performance fluctuations, with a higher total transmission rate than the OLLA-CMAB scheme but lower than the proposed scheme and the L-DQN scheme.
Furthermore, the proposed scheme demonstrates faster convergence and smaller performance fluctuations compared to the L-DQN scheme. This can be attributed to the fact that the proposed method leverages an GEXP-BO optimization module in the training phase, which benefits from the rapid convergence and strong stability of the BO algorithm.
This module provides reliable target actions in the training process, reducing estimation errors in policy gradient learning and thus accelerating the convergence rate. Additionally, the dual replay buffer sample selection strategy periodically trains the model with NACK samples, enabling the model to effectively handle NACK samples and thereby reducing performance fluctuations during training.

\begin{figure}[t!]
    \centering
    \includegraphics[width=.9\linewidth, trim = 0 0 0 10]{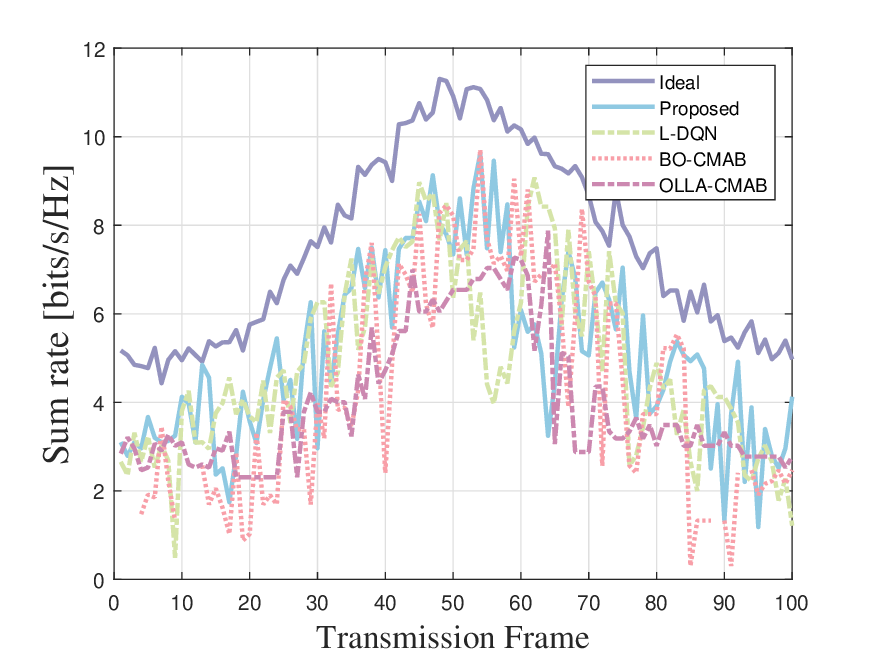}
    \caption{Test performance of different schemes versus communication round.}
    \label{Fig:Testing_performance}
\end{figure}
\begin{table}[t]
  \centering
  \caption{Average Test performance of different schemes}
  \begin{tabular}{cccc}
    \toprule
    & \textbf{Sum rate} & \textbf{BLER} & \textbf{Exceeded time} \\
    \midrule
    Ideal & 7.7153 & 0.001 & - \\
    Proposed & 6.1259 & 0.0076 & 26 \\
    L-DQN & 4.9257 & 0.0249 & 33 \\
    BO-CMAB & 4.0574 & 0.0551 & 49 \\
    OLLA-CMAB & 3.0188 & 0.1401 & 85 \\
    \bottomrule
  \end{tabular}
  \label{tab:average_performance}
\end{table}
In Fig.\ref{Fig:Testing_performance} and Table \ref{tab:average_performance}, we present performance evaluations for sum rate and BLER.  
In Table~\ref{tab:average_performance}, the average sum rate and BLER are calculated as $\frac{1}{K}\sum_{k=1}^{K}\hat{r}k$ and $\frac{1}{K}\sum_{k=1}^{K}\epsilon_k$, respectively, while ``exceeded time" indicates the total number of slots where BLER surpasses the thresholds. 
As observed in Fig.~\ref{Fig:Testing_performance}, the sum rate of the proposed algorithm closely approaches that of the ideal scheme, demonstrating adaptive and effective selection of device indices and MCS values under dynamic channel conditions. 
Moreover, the proposed scheme shows significantly lower performance fluctuations compared to the L-DQN and BO-CMAB schemes. Results in Table~\ref{tab:average_performance} further confirm the proposed method’s superior reliability, with the lowest average BLER and minimal threshold exceedances.
Conversely, although the OLLA-CMAB scheme exhibits stable performance fluctuations in Fig.~\ref{Fig:Testing_performance}, its overall performance is limited due to reliance on outdated CQI and ACK/NACK feedback, impeding real-time adjustments of MCS parameters. 
This reliance results in suboptimal MCS selections misaligned with actual channel conditions, thus reducing overall transmission efficiency.

\subsection{Wireless Parameter Evaluation}\label{Sec:sim_wireless_par}
\begin{figure}[htbp]
    \centering
    \includegraphics[width=0.9\linewidth, trim = 0 0 0 10]{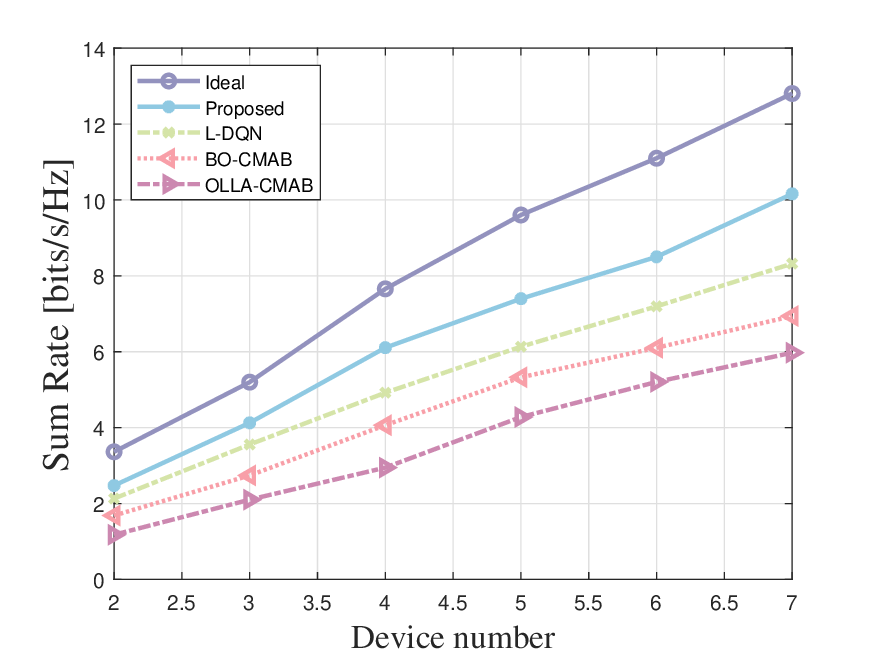}
    \caption{Sum rate performance of different schemes versus RD number.}
    \label{Fig:RD_num}
\end{figure}
\begin{figure}[htbp]
    \centering
    \includegraphics[width=0.9\linewidth, trim = 0 0 0 10]{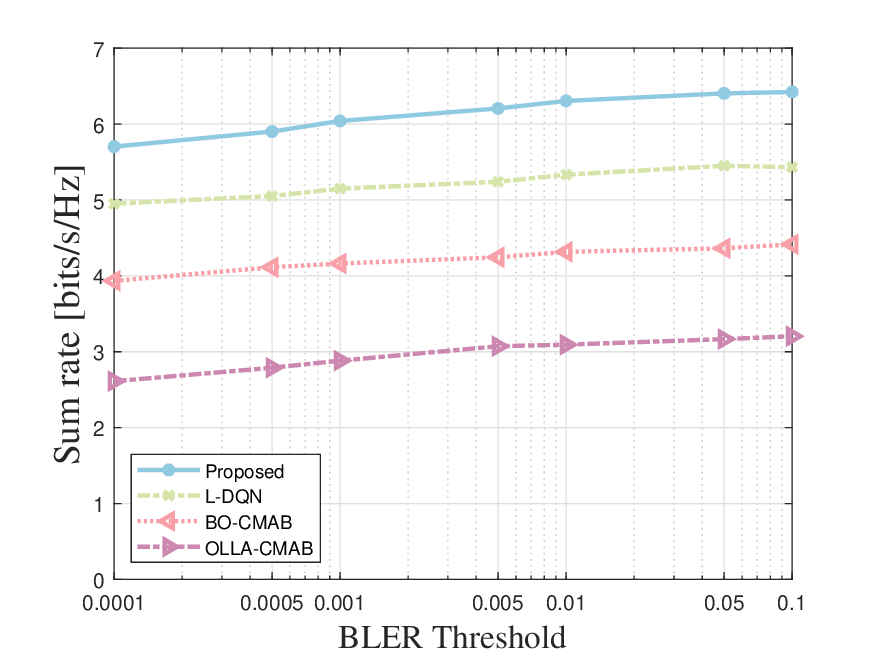}
    \caption{Sum rate performance of different algorithms versus BLER threshold.}
    \label{Fig:bler_threshold}
\end{figure}
Fig.~\ref{Fig:RD_num} illustrates the sum rate performance of different schemes with different numbers of devices, with each data point averaged over $30$ Monte Carlo simulations. 
It can be observed that the sum rate increases with the device count, and the proposed scheme consistently outperforms benchmark schemes. 
Notably, the absolute performance gap between the proposed and ideal schemes expands with more devices, from $0.893\ \text{bits/s/Hz}$ at 2 devices to $2.65\ \text{bits/s/Hz}$ at 7 devices. 
This phenomenon arises due to the elevated performance baseline and increased optimization complexity resulting from additional variables and channel dynamics. 
However, the relative performance gap diminishes from $26.6\%$ to $20.07\%$, confirming the scalability and effectiveness of the proposed scheme in multi-device networks.

Fig.~\ref{Fig:bler_threshold} illustrates the sum rate performance versus the BLER threshold for various schemes. 
As observed, increasing the BLER threshold enhances the sum rate across all methods, with the proposed scheme consistently achieving superior performance. 
This underscores the adaptability of our approach to diverse reliability requirements. 
Furthermore, raising the BLER threshold from $0.0001$ to $0.01$ yields a sum rate improvement of less than $10\%$, with diminishing returns observed beyond $0.01$. 
This trend highlights that the primary challenge in LA lies in effectively tracking dynamic channel variations. 
Although higher BLER thresholds expand the feasible MCS selection space and reduce optimization complexity, the resulting performance gains remain limited.

\subsection{Practical Discrete MCS network Evaluation}
\begin{figure}[t!]
    \centering
    \includegraphics[width=.8\linewidth, trim = 0 0 0 10]{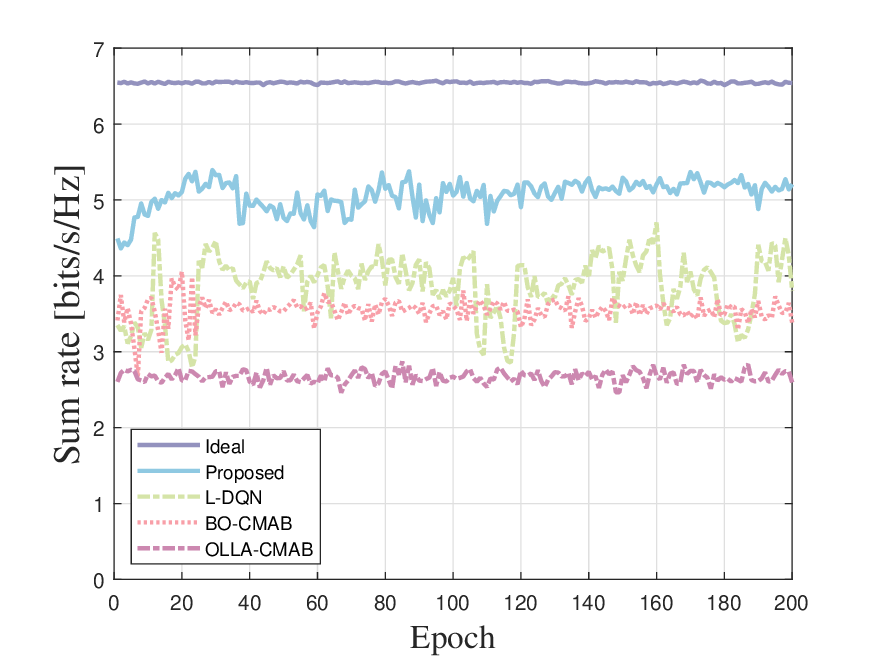}
    \caption{Sum rate performance of different schemes over training epoch with discrete MCS.}
    \label{Fig:Training_SP_quantized}
\end{figure}
The aforementioned simulations assume ideal conditions, modeling MCS (coding rate) as a continuous variable. 
However, practical URLLC networks impose discrete MCS constraints due to hardware and standard specifications. 
Thus, we further evaluate the proposed scheme's performance using discrete MCS indices defined by the 3GPP standard~\cite{CQI_MCS2}. 
Specifically, the MCS indices range from $8$ to $24$ as detailed in~\cite[Table 5.1.3.1-3]{MCS}. 
Notably, discretizing the problem transforms $(\text{OP})$ into an NP-hard problem, significantly increasing its complexity. 
Consequently, outputs from the proposed BO-assisted TD3 and baseline algorithms are rounded to the nearest discrete MCS index. 
Fig.~\ref{Fig:Training_SP_quantized} compares the sum-rate performances of various schemes under discretized MCS constraints. 
The results indicate that although discretization reduces performance for all methods due to limited adaptability to dynamic channel conditions, the proposed algorithm consistently outperforms benchmark schemes, validating its robustness in practical URLLC networks.
Upon discretizing the MCS values, the optimization formulation transitions from a mixed-integer programming problem to a pure integer programming problem, thereby reducing its complexity and accelerating convergence of the proposed scheme.

\begin{figure}[htbp]
    \centering
    \includegraphics[width=0.9\linewidth, trim = 0 0 0 10]{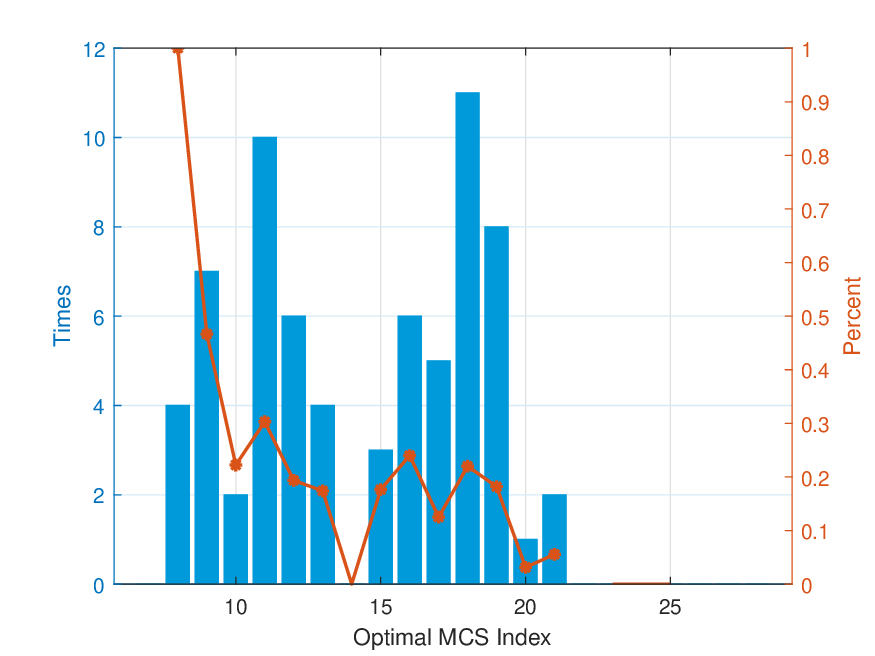}
    \caption{Percent versus MCS index.}
    \label{Fig:percent}
\end{figure}

\begin{figure}[htbp]
    \centering
    \includegraphics[width=0.9\linewidth, trim = 0 0 0 10]{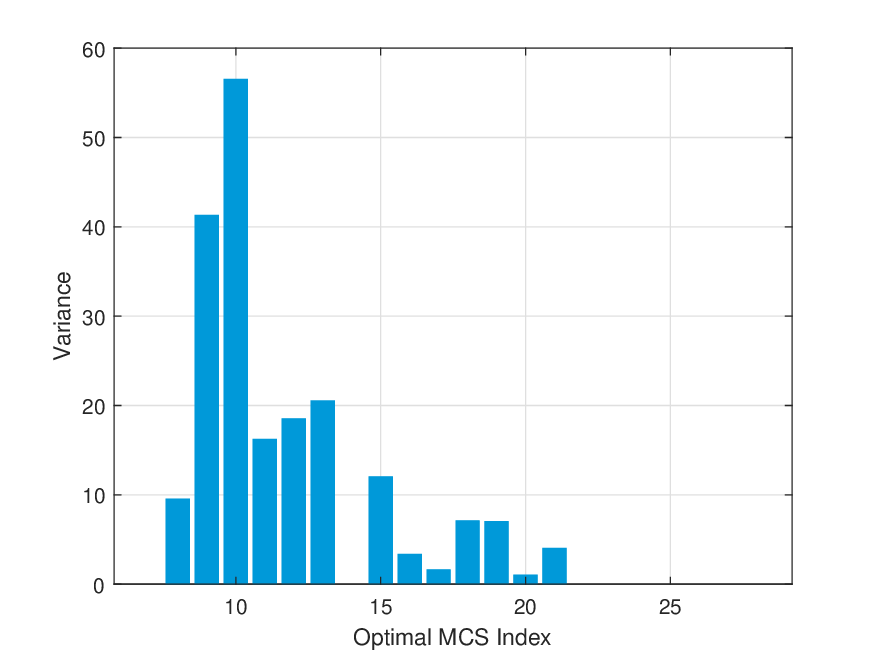}
    \caption{Exceeding the reliability Analysis.}
    \label{Fig:variance}
\end{figure}

Figs.\ref{Fig:Training_SP} to \ref{Fig:bler_threshold} presented comparisons of training and testing performance with baseline schemes. 
Next, we further assess the reliability of the proposed algorithm under events exceeding BLER threshold due to channel variability. 
Specifically, Fig.~\ref{Fig:percent} shows the frequency and percentage of reliability threshold violations for each optimal MCS index. 
The horizontal axis represents the optimal MCS index, while the left and right vertical axes denote, respectively, the absolute number and corresponding percentage of instances exceeding BLER thresholds. 
For instance, if the optimal MCS index is $15$ in $100$ transmissions out of $1000$ slots, with $10$ threshold violations, the values on the horizontal, left vertical, and right vertical axes would be $15$, $10$, and $10\%$, respectively. 
As illustrated, the violation rate remains below $25\%$ for optimal MCS indices above $13$, and notably drops below $10\%$ when indices exceed $22$. This reduction occurs because higher optimal MCS indices enlarge the feasible selection space, thus facilitating compliance with reliability constraints. Conversely, a smaller optimal MCS index narrows the selection space, increasing the violation frequency. Additionally, when the optimal index equals $8$, the total occurrences are very limited, causing even a small number of violations (four) to translate into a high violation percentage ($100\%$).

As illustrated in Fig.~\ref{Fig:variance}, when the optimal MCS index surpasses 16, the variance of the BLER exceeding the threshold is minimal, indicating that the selected MCS indices closely align with the optimal values, with only minor deviations. 
In contrast, when the optimal MCS index falls below $14$, the variance markedly increases, indicating substantial discrepancies between the selected and optimal MCS indices. 
This observation suggests that a low optimal MCS index constrains the MCS selection space (e.g., for an optimal MCS index of $9$, only indices $8$ and $9$ satisfy the BLER constraint). 
Consequently, in our future work, the proposed scheme’s performance under such circumstances remains a need for further improvement. 
Moreover, when the optimal MCS index surpasses $21$, the variance drops to $0$, indicating that all transmissions satisfy the BLER constraint.

\section{Conclusion}
In this work, we investigate a TDMA-based scheduling and MCS selection problem in dynamic URLLC networks characterized by imperfect CSI. 
To maximize the sum coding rate under strict BLER constraints, we propose a BO assisted TD3 framework to adaptively determine device scheduling order and corresponding MCS selections. 
To address prolonged convergence arising from DRL parameter sensitivity, we integrate a BO guided module into the training phase, enhancing action-selection reliability and accelerating convergence. 
Additionally, we introduce a sample selection mechanism to mitigate sample imbalance between ACK and NACK instances, further expediting training. Finally, to ensure compliance with stringent reliability requirements, an OLLA-assisted execution phase is incorporated, utilizing statistical channel information to refine the TD3's selected MCS values, thus enhancing overall network robustness.

Via simulation, we first demonstrate that our proposed algorithm outperforms baseline schemes in idealized networks, where the coding rate can be viewed as a continuous variable. Furthermore, the simulation results show that the proposed scheme surpasses benchmark approaches in both achieving a higher sum rate and minimizing instances where the BLER requirements are violated. 
These findings show that leveraging the strengths of artificial intelligence in addressing complex problems, in conjunction with the inherent stability of traditional mathematical analytical methods, can effectively realize complementary synergies, thereby significantly improving network performance.
Next, we further evaluate the proposed algorithm in a more realistic network, where the coding rate (or MCS)\pzr{ can only be chosen within}{ is quantized to a} certain discrete level\pza{s} according to the 3GPP standard. 
In this network, the proposed algorithm still outperforms baseline schemes. 
However, the simulation results show that the proposed algorithm achieves good performance when the optimal MCS index is large, while existed substantial discrepancies between the selected and optimal MCS indices
when the optimal MCS index is small. 
This motivated us to conduct further investigations under such circumstances for reliability improvement in our future work.

\bibliographystyle{IEEEtran}
\bibliography{reference}

\end{document}